\pdfoutput=1

\documentclass{article}

\usepackage{microtype}
\usepackage{graphicx}
\usepackage{booktabs} %
\usepackage{multirow}

\usepackage{bibentry}

\usepackage{amsmath}
\usepackage{amssymb}

\usepackage{siunitx}

\usepackage[textsize=tiny]{todonotes}
\usepackage{url}

\usepackage{enumitem}

\usepackage{caption}
\usepackage{subcaption}
\usepackage[normalem]{ulem}
\usepackage{wrapfig}
\usepackage{sidecap}

\usepackage{xspace}
\usepackage{tikz}
\usetikzlibrary{bayesnet}
\usetikzlibrary{arrows}
\usetikzlibrary{shapes.multipart}
\tikzset{
state/.style={
       rectangle split,
       rectangle split parts=2,
       rectangle split part fill={red!30,blue!20},
       rounded corners,
       draw=black, very thick,
       minimum height=2em,
       text width=3cm,
       inner sep=2pt,
       text centered,
       }
}
\usepackage{xcolor}
\definecolor{Blue}{rgb}{0.0,0.0,1.0}
\definecolor{Red}{rgb}{1.0,0.0,0.0}
\newcommand{\mf}[1]{}

\usetikzlibrary{shapes}
\usetikzlibrary{fit}
\usetikzlibrary{chains}
\usetikzlibrary{arrows}

\tikzstyle{latent} = [circle,fill=white,draw=black,inner sep=1pt,
minimum size=20pt, font=\fontsize{10}{10}\selectfont, node distance=1]
\tikzstyle{obs} = [latent,fill=gray!25]
\tikzstyle{const} = [rectangle, inner sep=0pt, node distance=1]
\tikzstyle{factor} = [rectangle, fill=black,minimum size=5pt, inner
sep=0pt, node distance=0.4]
\tikzstyle{det} = [latent, diamond]

\tikzstyle{plate} = [draw, rectangle, rounded corners, fit=#1]
\tikzstyle{wrap} = [inner sep=0pt, fit=#1]
\tikzstyle{gate} = [draw, rectangle, dashed, fit=#1]

\tikzstyle{caption} = [font=\footnotesize, node distance=0] %
\tikzstyle{plate caption} = [caption, node distance=0, inner sep=0pt,
below left=5pt and 0pt of #1.south east] %
\tikzstyle{factor caption} = [caption] %
\tikzstyle{every label} += [caption] %

\tikzset{>={triangle 45}}

\renewcommand{\edge}[3][]{ %
  \foreach \x in {#2} { %
    \foreach \y in {#3} { %
      \path (\x) edge [->,#1] (\y) ;%
    } ;
  } ;
}

\def\abovestrut#1{\rule[0in]{0in}{#1}\ignorespaces}

\usepackage{hyperref}

\usepackage[accepted]{icml2021}

\icmltitlerunning{Hierarchical VAEs Know What They Don't Know}

\newcommand{\J}{\mathbf{J}}
\newcommand{\y}{\mathbf{y}}
\newcommand{\x}{\mathbf{x}}
\newcommand{\z}{\mathbf{z}}
\newcommand{\I}{\mathbf{I}}
\newcommand{\0}{\mathbf{0}}

\setitemize{noitemsep,topsep=0pt,parsep=2pt,partopsep=0pt}

\begin{document}

\twocolumn[
\icmltitle{Hierarchical VAEs Know What They Don't Know}

\icmlsetsymbol{equal}{*}

\begin{icmlauthorlist}
\icmlauthor{Jakob D. Havtorn}{dtu,corti}
\icmlauthor{Jes Frellsen}{dtu}
\icmlauthor{S{\o}ren Hauberg}{dtu}
\icmlauthor{Lars Maal{\o}e}{dtu,corti}
\end{icmlauthorlist}

\icmlaffiliation{dtu}{Department of Applied Mathematics and Computer Science, Technical University of Denmark, Kongens Lyngby, Denmark}
\icmlaffiliation{corti}{Corti AI, Copenhagen, Denmark}

\icmlcorrespondingauthor{Jakob D. Havtorn}{jdh@corti.ai}
\icmlcorrespondingauthor{Lars Maal{\o}e}{lm@corti.ai}

\icmlkeywords{ICML, Machine learning, Generative modelling, Variational autoencoders, Out-of-distribution detection, Anomaly detection, AI safety}

\vskip 0.3in
]

\printAffiliationsAndNotice{}  %

\begin{abstract}
Deep generative models have been demonstrated as state-of-the-art density estimators.
Yet, recent work has found that they often assign a higher likelihood to data from outside the training distribution.
This seemingly paradoxical behavior has caused concerns over the quality of the attained density estimates.
In the context of hierarchical variational autoencoders, we provide evidence to explain this behavior by out-of-distribution data having in-distribution low-level features.
We argue that this is both expected and desirable behavior.
With this insight in hand, we develop a fast, scalable and fully unsupervised likelihood-ratio score for OOD detection that requires data to be in-distribution across all feature-levels.
We benchmark the method on a vast set of data and model combinations and achieve state-of-the-art results on out-of-distribution detection.
\end{abstract}

\section{Introduction}
\begin{figure}[t]
    \centering
    \includegraphics[width=1\columnwidth]{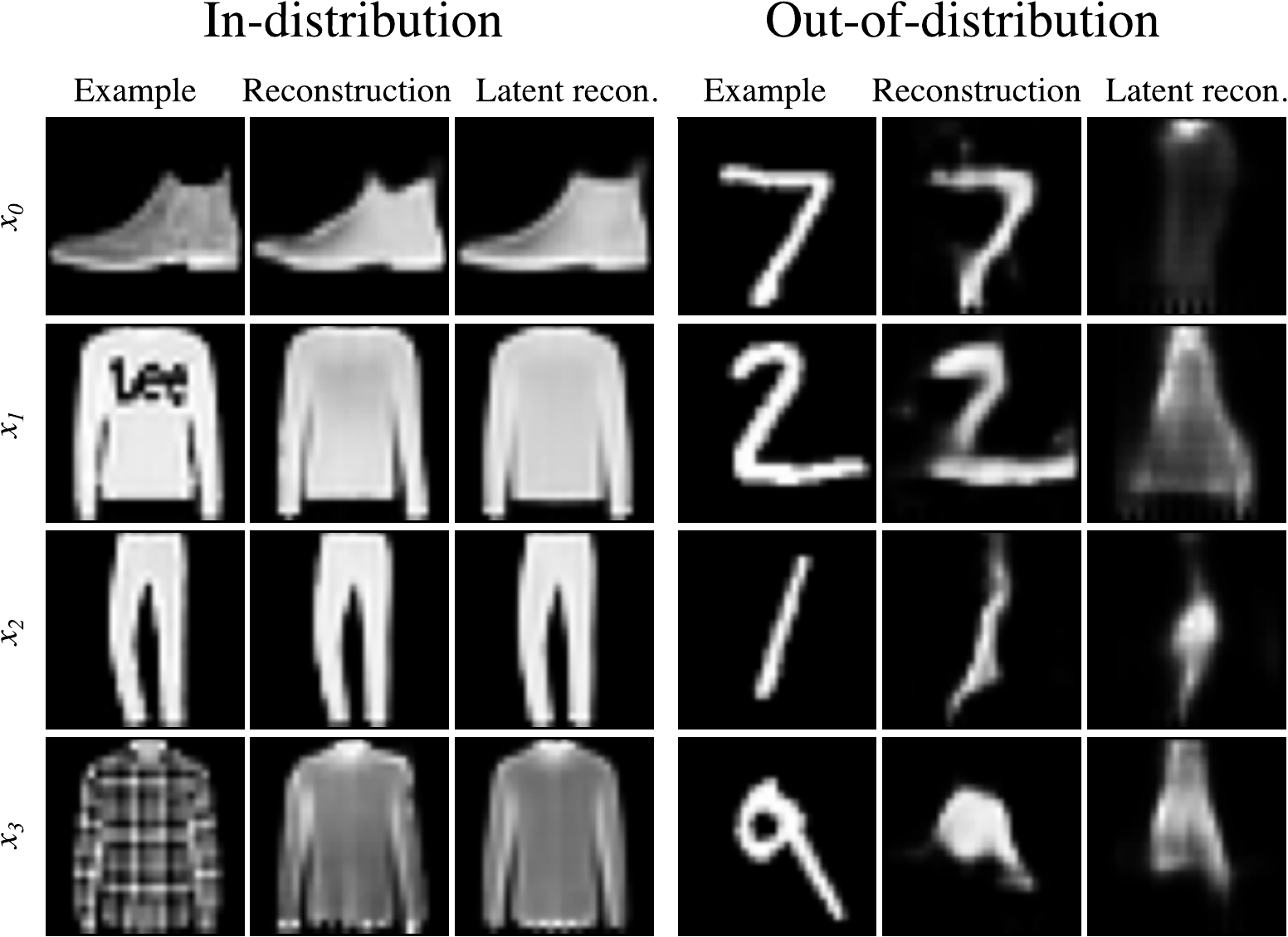}
    \vspace{-5mm}
    \caption{
        Reconstructions using a hierarchical VAE trained on FashionMNIST.
        Reconstruction quality of OOD data is comparable to in-distribution data, resulting in high likelihoods and poor OOD discrimination.
        By sampling the $k$ bottom-most latent variables from the conditional prior distribution $p(\z_{\geq l}|\z_{>l})$ (latent reconstructions) instead of the approximate posterior $q(\z_{>l}|\z_{<l})$, the model reconstructs from the training distribution resulting in lower $p(\x|\z)$ for OOD data.
    }
    \vspace{-5mm}
    \label{fig:reconstructions-fashionmnist}
\end{figure}
The reliability and safety of machine learning systems applied in the real-world is contingent on the ability to detect when an input is different from the training distribution. %
Supervised classifiers built as deep neural networks are well-known to misclassify such \textit{out-of-distribution} (OOD) inputs to known classes with high confidence \cite{goodfellow_explaining_2015, nguyen_deep_2015}.
Several approaches have been suggested to equip deep classifiers with OOD detection capabilities \cite{hendrycks_baseline_2017, lakshminarayanan_simple_2017, hendrycks_deep_2019, devries_learning_2018}.
But, such methods are inherently supervised and require in-distribution labels or examples of OOD data limiting their applicability and generality.

Unsupervised generative models that estimate an explicit likelihood should understand what it means to be in- and out-of-distribution without requiring labels or examples of OOD data.
By directly modeling the training distribution, such models are expected to assign low likelihoods to OOD data as it originates from regions of little or no support under the learned density \cite{bishop_novelty_1994}.
Recent advances in deep generative models \cite{kingma_auto-encoding_2014, rezende_stochastic_2014, oord_pixel_2016, salimans_pixelcnn_2017, kingma_glow_2018} have enabled learning high quality generative models on complex data such as natural images, sequences including audio \cite{oord_wavenet_2016} and graphs \cite{kipf_variational_2016}.
However, recent observations have brought into question the quality of the learned density estimates by showing that they often assign higher likelihoods to OOD data than to in-distribution data \cite{nalisnick_deep_2019, choi_waic_2019}.
Many complex data distributions can be explained to a large degree by low-level features, e.g. edges in images.
However, such features do not explain high-level semantics of the data and may inhibit OOD detection \cite{ren_likelihood_2019, nalisnick_deep_2019}

\textbf{In this paper}, we examine the failure cases of deep generative models on OOD detection tasks within the context of hierarchical VAEs, and make the following contributions:
\begin{itemize}
    \item[(i)] We provide evidence that the root cause of OOD failures is that learned low-level features generalize well across datasets and dominate the estimated likelihoods.
    \item[(ii)] We then propose a fast, scalable, and fully unsupervised likelihood-ratio score for OOD detection that is explicitly developed to ensure that data should be in-distribution across all feature levels, which prevents the low-level features from dominating.
    \item[(iii)] With the likelihood-ratio score, we demonstrate state-of-the-art performance across a wide range of known OOD failure cases.
\end{itemize}

\section{Why does OOD detection fail?}\label{sec:why-does-ood-fail}
\begin{figure}[t]
    \centering
    \includegraphics[width=1\columnwidth]{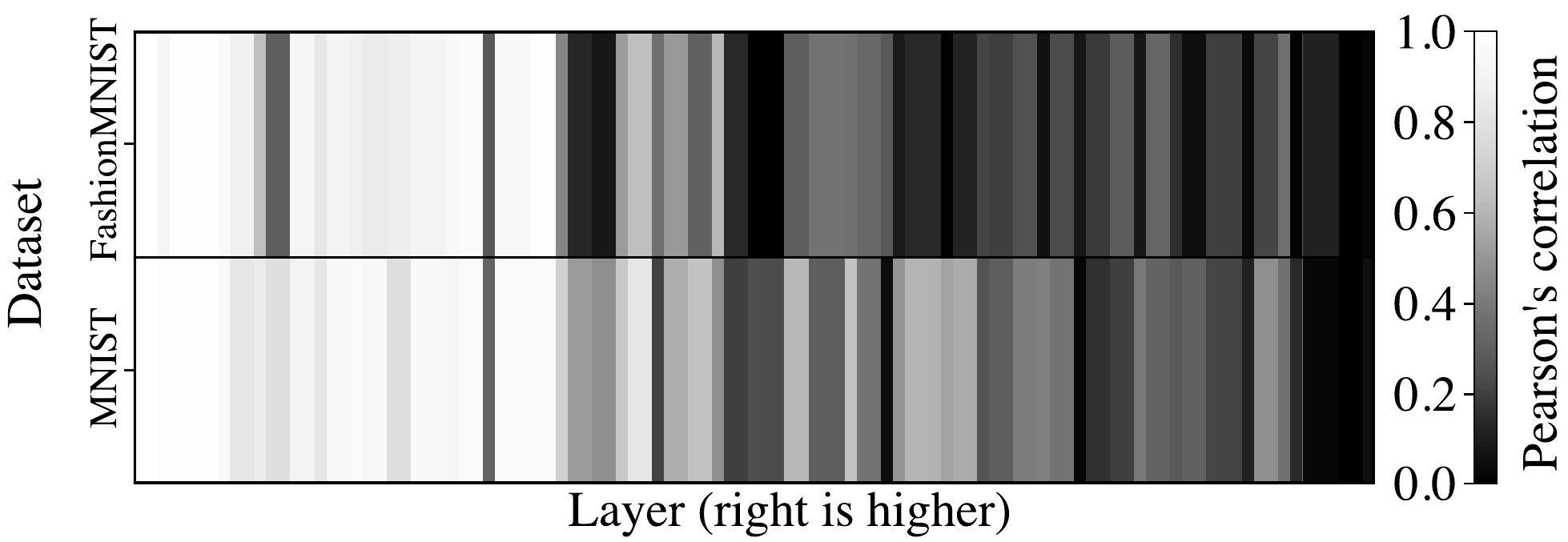}
    \vspace{-5mm}
    \caption{
    Absolute correlations between data representations in all layers of the inference network of a hierarchical VAE trained on FashionMNIST and of another trained on MNIST.
    We compute the correlation between the representations of the two different models given the same data, FashionMNIST (top) and MNIST (bottom).
    }
    \vspace{-3mm}
    \label{fig:correlations-heatmap}
\end{figure}

The inability to detect out-of-distribution data with deep generative models is surprising.
Before the advent of deep generative models, this was not considered a major issue for probabilistic models \cite{bishop_novelty_1994}.
Is the failure due to model pathologies or something different?

Deep learning models are generally believed to form hierarchies of representations that range from low-level features to more conceptual ones related to semantics \citep{bengio_representation_2013}.
This has also been observed within deep generative models \citep{maaloe_biva_2019, child_very_2021}.
For image data there is a trend that the low-level features are quite similar across models (edge detectors, etc.). This raises the question to what extend such features are relevant when detecting OOD data, also suggested  by \cite{nalisnick_deep_2019} and examined for Glow and PixelCNN in \cite{schirrmeister_understanding_2020}.
To investigate, we train two hierarchical VAEs (\autoref{sec:background-hie-VAE}) on FashionMNIST and MNIST, respectively, and compute the between-models correlation of the extracted features of in-distribution data and OOD data.
The result appears in \autoref{fig:correlations-heatmap}.
We observe that features extracted in the early layers (low-level features) correlate strongly between the two models, and that this correlation drops as we get into later layers.
This suggests that low-level features do not carry much information for OOD detection.

\begin{figure}[t]
    \centering
    \includegraphics[width=1\columnwidth]{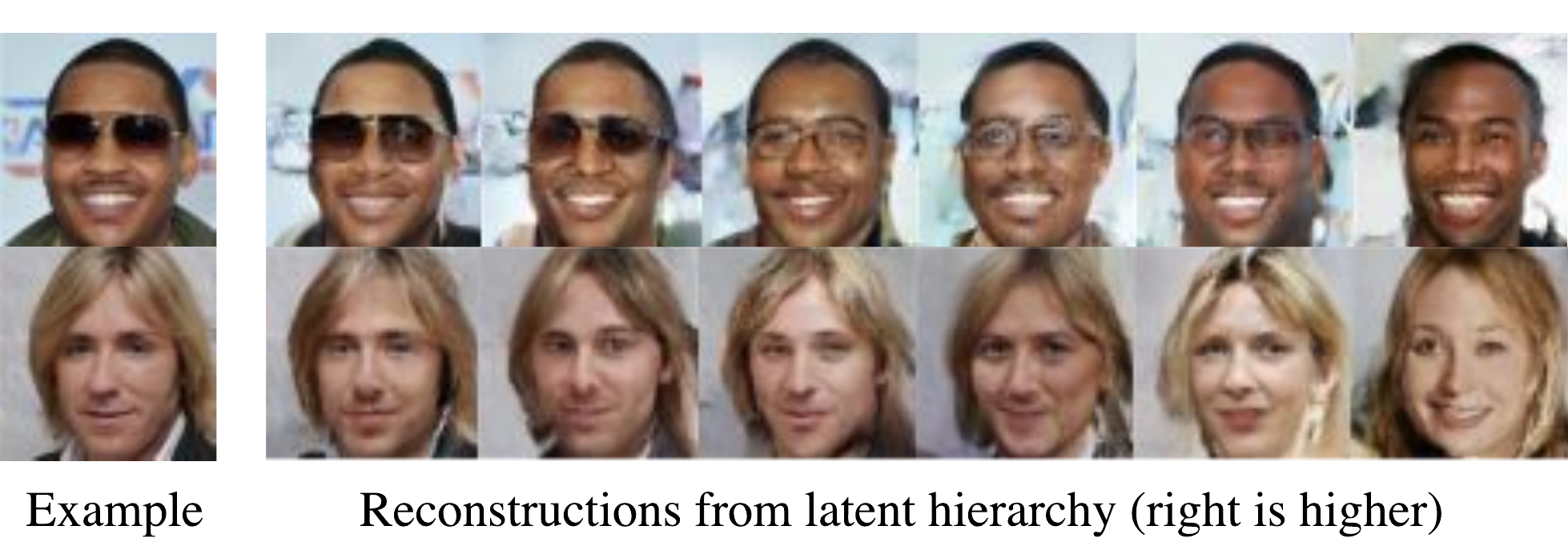}
    \vspace{-5.2mm}
    \caption{Reconstructions of in-distribution data (CelebA) of the BIVA model using higher latent variables  \cite{maaloe_biva_2019}.
    The higher the latent variable, the more the reconstructions fall into the mode of the learned distribution.
    It is more common to wear regular glasses than sunglasses but most common not to wear glasses at all.
    A man with long hair collapses into the mode of the more common long-haired woman.}
    \vspace{-3mm}
    \label{fig:biva-reconstructions-celeba}
\end{figure}

To shed further light on the impact of semantic versus low-level features, we look at model reconstructions of images with a hierarchical VAE (\autoref{fig:biva-reconstructions-celeba}).
To study the feature hierarchy, we replace the inference distribution with the corresponding conditional prior in the first layers of the model to see what information is lost.
We observe that as more layers rely on the prior, more details are lost.
Sunglasses, which are uncommon, are first replaced by more common glasses, and then finally disappear.
This suggests that as we fall back to the conditional priors of each layer, we are pushed closer to local modes of the modeled distribution.

Finally, we look at reconstructions of out-of-distribution data.
\autoref{fig:reconstructions-fashionmnist} illustrates that MNIST data is surprisingly well reconstructed by a hierarchical VAE trained on FashionMNIST.
Similar results have been found elsewhere \cite{xiao_likelihood_2020}.
We repeat the previous experiment and replace inference distributions by their corresponding conditional prior, and now observe that reconstructions from higher latent layers become increasingly similar to the data on which the model was trained.
The reliance on conditional priors seems to prevent accurate reconstruction of out-of-distribution data.
Some details are lost on in-distribution data too, but the distinction between that and out-of-distribution data becomes more clear.

\textbf{These observations lead to our main hypothesis.}
The lowest latent variables in a hierarchical VAE learn generic features that can describe a wide range of data.
This enables the model to achieve high rates of compression and high likelihoods, even on out-of-distribution data as long as the learned low-level features are appropriate.
We further suggest that OOD data are in-distribution with respect to these low-level features, but not with respect to semantic ones.

\vspace{-0.35cm}
\section{Background and related work}

\subsection{Variational autoencoders}
The variational autoencoder (VAE) \cite{kingma_auto-encoding_2014, rezende_stochastic_2014} is a framework for constructing deep generative models defined by an observed variable $\mathbf{x}$ and a stochastic latent variable $\mathbf{z}$.
Typically, a neural network with parameters $\theta$ is chosen to parameterize the generative distribution $p_\theta(\x,\z)=p_\theta(\x|\z)p(\z)$, where the prior $p(\z)$ is commonly a standard Gaussian $\mathcal{N}(\0, \I)$.
The true posterior $p(\z|\x)$ is generally not analytically tractable and is approximated by a variational distribution $q_\phi(\z|\x)$ parameterized via another neural network with parameters $\phi$. The approximate posterior $q_\phi(\z|\x)$ is most often  a diagonal covariance Gaussian.
The model parameters $\theta$ and variational parameters $\phi$ are jointly optimized by maximizing the \textit{evidence lower bound} (ELBO),
\begin{equation}\label{eq:elbo}
    \log p_\theta(\x) \geq \mathbb{E}_{q_\phi(\z|\x)} \left[ \log \frac{p_\theta(\x,\z)}{q_\phi(\z|\x)} \right] \equiv \mathcal{L}(\x; \theta, \phi)\ .
\end{equation}
For brevity, we will denote $\mathcal{L}(\x; \theta, \phi)$ as $\mathcal{L}(\x)$ or $\mathcal{L}$. The reparameterization trick is used to backpropagate gradients through the stochastic latent variables with low variance.

The VAE is defined with a single latent variable which limits the ability to learn a high likelihood representation of complex input distributions, e.g.\ natural images.
There exists a few complementary approaches to make the VAE more flexible: (i) model a more expressive variational distribution $q_\phi(\z|\x)$ or prior distribution $p_\theta(\z)$ \cite{rezende_variational_2015, kingma_improved_2016}, (ii) model a more expressive posterior distribution $p_\theta(\x|\z)$ e.g. with a autoregressive decoder \cite{van_den_oord_conditional_2016} and (iii) learn a deeper hierarchy of latent variables \cite{burda_importance_2016, sonderby_ladder_2016}.
Here, we focus on the latter.

\subsection{Hierarchical variational autoencoders}\label{sec:background-hie-VAE}
Hierarchical VAEs are a family of probabilistic latent variable models which extends the basic VAE by introducing a hierarchy of $L$ latent variables $\z=\z_1, \dots, \z_L$.
The most common generative model is defined from the top down as $p_\theta(\x|\z)=p(\x|\z_1)p_\theta(\z_1|\z_2)\cdots p_\theta(\z_{L-1}|\z_L)$.
The inference model can then be defined in two ways respectively referred to as \textit{bottom-up} \cite{burda_importance_2016}
\begin{equation}
    q_\phi(\z|\x) = q_\phi(\z_1|\x)\textstyle\prod_{i=2}^{L} q_\phi(\z_i|\z_{i-1})
\end{equation}
and \textit{top-down} \cite{sonderby_ladder_2016}
\begin{equation}
    q_\phi(\z|\x) = q_\phi(\z_L|\x)\textstyle\prod_{i=L-1}^{1} q_\phi(\z_{i}|\z_{i+1}) \ .
\end{equation}
Regardless of the choice of inference model, a hierarchical VAE is still trained using the ELBO \eqref{eq:elbo}.

Until recently, hierarchical VAEs gave inferior likelihoods compared to state-of-the-art autoregressive \cite{ho_flow_2019} and flow-based models \cite{salimans_pixelcnn_2017}.
This was changed by \citet{maaloe_biva_2019}, \citet{vahdat_nvae_2020}, and \citet{child_very_2021}, which introduced complementary methods to extend the number of latent variables to a very deep hierarchy resulting in state-of-the-art likelihood performance.

In this paper we employ a simple hierarchical VAE with bottom-up inference paths and the more powerful BIVA variant with a bidirectional (top-down and bottom-up) inference model \citep{maaloe_biva_2019}. We employ skip connections between latent variables but omit them for brevity.

\subsection{Out-of-distribution detection}\label{sec:background-ood-detection}
So far, no reliable direct likelihood-based method has been found for fully unsupervised deep generative model OOD detection.
A major line of work considers developing new scores that are more reliable than the likelihood.
This includes the \textit{typicality} test presented by \citet{nalisnick_detecting_2019} which is an OOD detection test based on the typicality of a batch of potentially OOD examples.
This approach however requires a batch of examples from the same class (OOD or not) which limits its practical applicability.
In \citet{ren_likelihood_2019}, the \textit{likelihood ratio} between a primary model and a background model was shown to be an effective score for OOD detection.
However, to train the background model, the in-distribution data is perturbed via a data augmentation technique that is designed with knowledge about the confounding factors between the in-distribution data and the OOD data. Furthermore, it is tuned towards high performance on a known OOD dataset.
\citet{serra_input_2020} take a similar approach and attribute the failure to detect OOD data to the high influence of the input complexity on the likelihood and choose a generic lossless compression algorithm as the background model.
Although this method gives good results, no single best choice of compression algorithm exists for all types of OOD data, and any particular choice encodes prior knowledge about the data into the detection method.
Both these methods can be seen as correcting for low-level features of the OOD data being assigned high model likelihood by using a second model focused exclusively on these features.

Similar to these methods, the majority of the approaches to OOD detection make assumptions about the nature of the OOD data.
The assumptions encompass using labels on the in-distribution data \cite{hendrycks_baseline_2017, liang_enhancing_2018, alemi_uncertainty_2018, lee_simple_2018, lakshminarayanan_simple_2017}, examples of OOD data \cite{hendrycks_deep_2019}, augmenting in-distribution data to mimic it \cite{ren_likelihood_2019}, or assuming a certain data type \cite{serra_input_2020}.
Any of these assumptions encode implicit biases into the model about the attributes of OOD data which, in turn, might impair performance on truly unknown data examples (unknown unknowns).

While some of these methods achieve very good results on OOD detection with autoregressive models \cite{oord_pixel_2016, salimans_pixelcnn_2017} and invertible flow-based models \cite{kingma_glow_2018}, it was recently shown that they can be much less effective for VAEs \cite{xiao_likelihood_2020} highlighting the need for a more reliable OOD score for VAEs.
Although VAEs have the same failure cases as autoregressive and flow-based models, the caveat is that the difference in the likelihood is generally not as big and reconstructions of OOD can be surprisingly good \cite{xiao_likelihood_2020}.
\citet{xiao_likelihood_2020} alleviate this by refitting the inference network, as previously proposed by \citet{cremer_inference_2018, mattei_refit_2018}, to a potentially OOD example and measuring the so-called \textit{likelihood regret}.
However, refitting the inference network can be computationally expensive, especially for the large hierarchical VAEs that are used to model complex data \cite{maaloe_biva_2019, vahdat_nvae_2020, child_very_2021}. Furthermore, this scales poorly to large amounts of potentially OOD examples as the optimization is done per example.

A few methods have approached OOD detection in a completely unsupervised fashion \cite{maaloe_biva_2019, choi_waic_2019, xiao_likelihood_2020}.
The work of \citet{maaloe_biva_2019} is the most related to ours. They introduce BIVA, a deep hierarchy of stochastic latent variables with a top-down and bottom-up inference model and achieve state-of-the-art likelihood scores. 
They also provide early results indicative that a looser likelihood bound may have value in OOD detection.
In this paper, we provide an explanation of those results, and significantly improve upon them.

\section{OOD detection with hierarchical VAEs}
\subsection{A bound for semantic OOD detection}
If the lowest latent variable in the VAE hierarchy codes for a large part of the low-level features required to reconstruct the input with high accuracy, as exemplified in \autoref{fig:reconstructions-fashionmnist}-\ref{fig:biva-reconstructions-celeba}, then $p_\theta(\x|\z_1)$ will be high for both in- and out-of-distribution data.
Hence, any OOD detection capabilities based on the ELBO $\mathcal{L} = \mathbb{E}_{q_\phi(\z|\x)}[\log p_\theta(\x|\z_1)] - D_{\mathrm{KL}}( q_\phi(\z|\x) || p(\z))$ from \eqref{eq:elbo} relies on the KL-term for OOD detection. For a bottom-up hierarchical VAE, the KL-term $D_{\mathrm{KL}}( q_\phi(\z|\x) || p(\z))$ can be expressed by a hierarchical sum%
\begin{equation}
    \mathbb{E}_{q_\phi(\z|\x)} \Big[ \textstyle\sum_{i=1}^{L-1} \log \frac{p_\theta(\z_i|\z_{i+1})}{q_\phi(\z_i|\z_{i-1})} + \log \frac{p_\theta(\z_L)}{q_\phi(\z_L|\z_{L-1})} \Big] \ .
\end{equation}
In general, the absolute log-ratios grow with $\mathrm{dim}(\z_i)$ as the individual log probability terms are computed by summing over the dimensionality of $\z_i$.
This means that the value of the KL-term is dominated by terms where $\z_i$ is high-dimensional. We refer to Appendix C for a more detailed argument.
Since hierarchical VAEs are generally constructed with a bottleneck type structure, the terms corresponding to latent variables towards the top of the hierarchy will have a vanishing influence on the value of the KL-term.
However, as the semantic information most relevant for OOD detection has a tendency to be represented in the top-most latent variables, this makes OOD detection using the regular ELBO difficult, even for state-of-the-art models.
This behavior has also been reported by \citet{xiao_likelihood_2020}.

To shift the ELBO from primarily being based on the approximate posterior of the lowest latent variables to instead focus on the conditional prior, \citet{maaloe_biva_2019} introduced slightly different likelihood lower bound defined as
\begin{equation}\label{eq:biva->k}
    \mathcal{L}^{>k} = \mathbb{E}_{p_\theta(\z_{\leq k}|\z_{>k}) q_\phi(\z_{>k}|\x)} \left[ \log \frac{p_\theta(\x|\z)p_\theta(\z_{>k})}{q_\phi(\z_{>k}|\x)} \right]
\end{equation}
where $k\in\{0,1,\dots,L\}$ (see Appendix for the derivation).
We note that $\mathcal{L}^{>0}$ is the regular ELBO (\ref{eq:elbo}) and that empirically we always observe that $\mathcal{L}\geq\mathcal{L}^{>k} \, \forall \, k$ although this need not hold in general.
The core idea behind this variation on the ELBO is to sample the $k$ lowest latent variables from the conditional prior $\z_1,\dots,\z_l \sim p_\theta(\z_{\leq k}|\z_{>k})$ and only the $L-k$ highest from the approximate posterior $\z_{k+1},\dots,\z_L \sim q_\phi(\z_{>k}|\x)$.
Importantly, this has the effect that the data likelihood $p(\x|\z)$ is dependent on the approximate posterior through a latent variable $\z_{k+1}$ different from $\z_1$ for all $k \geq 1$.
Thereby, the likelihood can be evaluated with a reconstruction from each of the latent variables $\z_k$ of the hierarchical VAE.
Hence, we can now test how well the input $\x$ is reconstructed from each latent variable.
The notation $\mathcal{L}^{>k}$ highlights that for latent variables $\z_{>k}$, the bound is the regular ELBO while for the latent variables $\z_{\leq k}$, the bound is evaluated using the (conditional) prior rather than the approximate posterior as the proposal distribution.

\subsection{A likelihood-ratio score for all feature levels}
While the $\mathcal{L}^{>k}$ bound provides a score for performing semantic OOD detection, it still relies on the data space likelihood function (see equation \eqref{eq:likelihoods-as-exact} below), which is known to be problematic for OOD detection (section \ref{sec:background-ood-detection}). To alleviate this, we phrase OOD detection as a likelihood ratio test of being \emph{semantically} in-distribution.
A standard likelihood ratio test \citep{buse_likelihood_1982} suggests to consider the ratio between the associated likelihoods, which we can approximate on a log-scale by the corresponding lower bounds $\mathcal{L}$ and $\mathcal{L}^{>k}$,
\begin{equation}\label{eq:llr-as-difference-in-likelihoods}
    LLR^{>k}(\x) = \mathcal{L}(\x) - \mathcal{L}^{>k}(\x) \ .
\end{equation}
Since, empirically, $\mathcal{L}\geq\mathcal{L}^{>k}$, the ratio is always positive as is standard for likelihood ratio tests.
A low value of $LLR^{>k}(\x)$ means that the ELBO and $\mathcal{L}^{>k}$ are almost equally tight for the data.
On the contrary, a high value indicates that $\mathcal{L}^{>k}$ is looser on the data than the ELBO; hence, the data may be OOD.

We can gather further insights about this score if we write the regular ELBO and the $\mathcal{L}^{>k}$ bounds in the exact form that includes the intractable KL-divergence between the approximate and true posteriors,
\begin{align}
    \mathcal{L}      &= \log p_\theta(\x) - D_{\mathrm{KL}}\left( q_\phi(\z|\x) || p_\theta(\z|\x)\right), \label{eq:likelihoods-as-exact} \\ 
    \mathcal{L}^{>k} &= \log p_\theta(\x) - D_{\mathrm{KL}}\left( p_\theta(\z_{\leq k }|\z_{>k}) q_\phi(\z_{>k}|\x) || p_\theta(\z|\x)\right) \nonumber \ .
\end{align}
Subtracting these cancel out the two data likelihood terms $\log p_\theta(\x)$ and only the KL-divergences from the approximate to the true posterior remain,
\begin{align}\label{eq:llr-as-kls}
    LLR^{>k}(\x) &= - D_{\mathrm{KL}}\left( q_\phi(\z|\x) || p_\theta(\z|\x)\right) \\
                 &\quad + D_{\mathrm{KL}}\left( p_\theta(\z_{\leq k}|\z_{>k}) q_\phi(\z_{>k}|\x) || p_\theta(\z|\x)\right) \ . \notag
\end{align}

Hence, it is clear that compared to the likelihood bound $\mathcal{L}^{>k}$, this likelihood-ratio measures divergence exclusively in the latent space whereas $\mathcal{L}^{>k}$ includes the $\log p_\theta(\x)$ term similar to the ELBO.
Therefore, the $LLR^{>k}$ score should be an improved method for semantic OOD detection compared to $\mathcal{L}^{>k}$.
Now, it can be noted that if we replace the regular ELBO, $\mathcal{L}$, in \eqref{eq:likelihoods-as-exact} with the strictly tighter importance weighted bound \cite{burda_importance_2016},
\begin{equation}\label{eq:iw-bound}
    \mathcal{L}_{S} = \mathbb{E}_{q(\z|\x)}\left[ \log \frac{1}{N} \sum_{s=1}^{S} \frac{p(\x, \z^{(s)})}{q(\z^{(s)}|\x)} \right] \ ,
\end{equation}
then, in the limit $S\rightarrow\infty$, we have $\mathcal{L}_{S} \rightarrow \log p_\theta(\x)$ and the likelihood ratio reduces to
\begin{equation}\label{eq:llr-as-kls-iwae-reduced}
    LLR^{>k}_{S}(\x) \rightarrow D_{\mathrm{KL}}( p(\z_{\leq k}|\z_{>k}) q(\z_{>k}|\x) || p(\z|\x))
\end{equation}
which, in practice, is well-approximated for a finite $S$. We expect this importance weighted likelihood ratio to monotonically improve upon the one in (\ref{eq:llr-as-kls}) as $S$ increases and the KL-divergence in the regular ELBO that contains terms for which $\z_i$ is high-dimensional goes to zero.

Since the scores in (\ref{eq:llr-as-kls}) and (\ref{eq:llr-as-kls-iwae-reduced}) are estimated by sampling their estimators are stochastic objects with nonzero variance.
We note that $\text{Var}(\widehat{LLR}^{>k}) = \text{Var}(\hat{\mathcal{L}}) + \text{Var}(\hat{\mathcal{L}}^{>k}) - 2\, \text{Cov}(\hat{\mathcal{L}}, \hat{\mathcal{L}}^{>k})$.
Since $\log p_\theta(\x)$ and part of the KL divergence are identical in the expressions of $\mathcal{L}$ and $\mathcal{L}^{>k}$ we expect $\text{Cov}(\hat{\mathcal{L}}, \hat{\mathcal{L}}^{>k})$ to be positive which reduces the total variance. 
Empirical results indeed show that $\text{Var}(\widehat{LLR}^{>k})$ is larger than $\text{Var}(\hat{\mathcal{L}})$ but smaller than $\text{Var}(\hat{\mathcal{L}}^{>k})$.
Nevertheless, the variance of the estimators is guaranteed to go to zero as the number of samples is increased.

The OOD scores considered in this research all assume that what discriminates an out-of-distribution from an in-distribution data point are semantic, high-level features. Clearly, if this is not the case and the difference instead lies in low-level statistics, the scores would likely fail. We hypothesize that a complementary bound to (\ref{eq:biva->k}), $\mathcal{L}^{<l}$ described in Appendix E, might be useful in these cases, but leave further examination to future work.

\section{Experimental setup}

\textbf{Tasks}: We follow existing literature \cite{nalisnick_deep_2019, hendrycks_deep_2019} and evaluate our method by setting up OOD detection tasks from FashionMNIST \cite{xiao_fashion-mnist_2017} to MNIST \cite{lecun_gradient-based_1998} and from CIFAR10 \cite{krizhevsky_learning_2009} to SVHN \cite{netzer_reading_2011}.
For each experiment we train our model on the train split of the former dataset and test its ability to recognize the test split of the latter dataset as OOD from the test split of the former dataset.
We use the standard train/test splits for the datasets.
More details on the datasets can be found in the Appendix.

\begin{SCfigure}[50][t!]
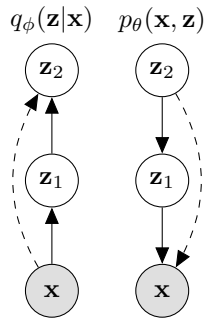

    \tikz{
        \node[obs] (x_inf) {$\x$};%
        \node[latent,above=.75cm of x_inf](z1_inf){$\z_1$}; %
        \node[latent,above=.75cm of z1_inf](z2_inf){$\z_2$}; %
        \node[above=of z2_inf, yshift=-1.cm] (phi) {$q_\phi(\z|\x)$}; 
        
        \edge[]{x_inf}{z1_inf};
        \edge[]{z1_inf}{z2_inf};
        \edge[dashed, bend left]{x_inf}{z2_inf};
        
        \node[obs,right=0.75cm of x_inf] (x_gen) {$\x$};%
        \node[latent,above=.75cm of x_gen](z1_gen){$\z_1$}; %
        \node[latent,above=.75cm of z1_gen](z2_gen){$\z_2$}; %
        \node[above=of z2_gen, yshift=-1.cm] (theta) {$p_\theta(\x,\z)$}; 
        
        \edge[]{z2_gen}{z1_gen};
        \edge[]{z1_gen}{x_gen};
        \edge[dashed, bend left]{z2_gen}{x_gen};
    }
    \vspace{-5mm}
    \caption{The inference and generative models, $q_\phi$ and $p_\theta$, for an $L=2$ layered bottom-up hierarchical VAE as the one used in our experiments.
    Dashed lines indicate deterministic skip connections which are employed in both networks. Skip connections are found to be useful for optimizing latent variable models \citep{dieng_avoiding_2019, maaloe_biva_2019}.}
    \label{fig:hvae-graphical-model}
    \vspace{5mm}
\end{SCfigure}

\textbf{Models}: For each OOD task, we train a simple bottom-up hierarchical VAE with $L$ stochastic layers which we will refer to as ``HVAE''.
To alleviate posterior collapse we include skip-connections that connect $\z_i$ to $\z_{i+2}$ for $i\in\{0, L-2\}$ and $\z_0\equiv\x$ in both the inference and generative models \cite{dieng_avoiding_2019} and employ the \textit{free bits} scheme with $\lambda=2$ \cite{kingma_improved_2016}.
We use weight-normalization \cite{salimans_weight_2016} on all weights and residual networks in the deterministic paths. 
A graphical representation of this model can be seen in \autoref{fig:hvae-graphical-model}.
We use a Bernoulli output distribution for FashionMNIST/MNIST and a discretized mixture of logistics output distribution \cite{salimans_pixelcnn_2017} for CIFAR10/SVHN.
We use $L=3$ for grey-scale images and $L=4$ for natural images.
For CIFAR/SVHN, we also train a BIVA model \cite{maaloe_biva_2019} with $L=10$ and similar configuration as used by the original paper\footnote{Source code available at \newline\url{github.com/larsmaaloee/BIVA} and \newline\url{github.com/vlievin/biva-pytorch}}.
All models are trained by optimizing the ELBO in (\ref{eq:elbo}). We implement our models in PyTorch \cite{paszke_automatic_2017}\footnote{Source code available at \newline\url{github.com/jakobhavtorn/hvae-oodd}}.
Full model details are in the Appendix.

\textbf{Baselines}: We group baselines into those that use prior knowledge about OOD data, ones that use labels associated with the in-distribution data and purely unsupervised approaches that do not make such assumptions.
Our method falls into the latter category.
For more information on each baseline, we refer to the original literature.

\textbf{Evaluation}: Following previous work \cite{hendrycks_baseline_2017, hendrycks_deep_2019, alemi_uncertainty_2018, ren_likelihood_2019, choi_waic_2019} we use the threshold-independent evaluation metrics of Area Under the Receiver Operator Characteristic (AUROC$\uparrow$), Area Under the Precision Recall Curve (AUPRC$\uparrow$) and False Positive Rate at 80\% true positive rate (FPR80$\downarrow$) where the arrow indicates the direction of improvement.
Note that these metrics are only computable given examples of OOD data but faced with truly OOD data (unknown unknowns), there are many ways to select thresholds to use in practice e.g.\ as the one that yields a specific tolerable false positive rate on the in-distribution test data.
To compute the metrics, we use an equal number of samples from the in-distribution and OOD datasets by including all examples in the smallest of the two sets and randomly sampling equally many from the larger. We compute the $LLR^{>k}$ score with one and $S$ importance samples denoted by $LLR^{>k}_S$.

\textbf{Selection of $k$}: To determine whether an example is OOD in practice, the value of $LLR^{>k}$ is computed on the in-distribution test set for all $k$ and the resulting empirical distribution is used as reference.
If for any value of $k$, the $LLR^{>k}$ score of a new input differs significantly from the empirical distribution, it is regarded OOD.
If it differs for multiple values of $k$, the value for which it differs the most is selected.
In our experiments, we consider an entire dataset at a time and report the results of $LLR^{>k}$ with the value of $k$ that yielded the highest AUROC$\uparrow$ for that dataset in a threshold-free manner.
In practice, slightly better performance may be achieved by choosing $k$ per example.
This would not exclude the use of batching in our method, since $LLR^{>k}$ is computed after the forward pass.

\section{Results}

The likelihoods for our trained models are in \autoref{tab:bits-per-dim-ood} alongside baseline results for in-distribution and OOD data.
The main results of the paper on the OOD tasks can be seen along with comparisons to the baseline methods in \autoref{tab:rocauc-ood}.
We note that for all our results, the value of the score ($\mathcal{L}^{>k}$ and $LLR^{>k}$) for the training and test splits of the in-distribution data was observed to have the same empirical distribution to within sampling error hence yielding an AUROC score of $\approx0.5$ as expected.
Results on additional commonly used datasets are found in Appendix G.

\begin{table}[t!]
    \centering
    \resizebox{\columnwidth}{!}{%
    \begin{tabular}{lrrrrr}
        \toprule
         Method & Dataset & \multicolumn{4}{c}{Avg. bits/dim}\\
          & & $\log p(x)$ & $\mathcal{L}^{>1}$ & $\mathcal{L}^{>2}$ & $\mathcal{L}^{>3}$\\
         \midrule
         \multicolumn{6}{c}{\textbf{Trained on FashionMNIST}} \\
         \midrule
         \multirow{2}{*}{Glow}
            & FashionMNIST & 2.96 & - & - & \\
            & MNIST & 1.83 & - & - & \\
         \multirow{2}{*}{HVAE (Ours)}
            & FashionMNIST & 0.420 & 0.476 & 0.579 & - \\
            & MNIST & 0.317 & 0.601 & 0.881 & - \\
         \midrule
         \multicolumn{6}{c}{\textbf{Trained on CIFAR10}} \\
         \midrule
         \multirow{2}{*}{Glow}
          & CIFAR10 & 3.46 & - & - & \\
          & SVHN & 2.39 & - & - & \\
         \multirow{2}{*}{HVAE (Ours)}
            & CIFAR10 & 3.74 & 17.8 & 54.3 & 75.7 \\  %
            & SVHN & 2.62 & 10.2 & 64.0 & 93.9 \\
         \multirow{2}{*}{BIVA (Ours)}
          & CIFAR10 & 3.46 & 8.74 & 19.7 & 37.3 \\
          & SVHN & 2.35 & 6.62 & 25.1 & 59.0 \\
         \bottomrule
    \end{tabular}
    }
    \caption{
    Average bits per dimension of different datasets for models trained on FashionMNIST and CIFAR10.
    For the hierarchical models we include the $\mathcal{L}^{>k}$ bounds.
    The likelihoods of training and test splits of the in-distribution data are all cases close.
    Since we train on dynamically binarized FashionMNIST, our bits/dim are smaller than for Glow.
    As $k$ is increased for the $L^{>k}$ bound, the bound gets looser but the model eventually assigns higher likelihood to the in distribution data than to the OOD data.
    Glow refers to \citet{kingma_glow_2018, nalisnick_deep_2019}.
    BIVA refers to our implementation of \citet{maaloe_biva_2019}.}
    \label{tab:bits-per-dim-ood}
    \vspace{-3mm}
\end{table}

\footnotetext{
    \citet{serra_input_2020} performs the best when high likelihoods are assigned to OOD data such that the overlap with in-distribution data is low.
    Performance is worse when the overlap is high, cf. \citet[Table 1]{serra_input_2020}, as seen with complex images. %
}
\begin{table}[t!]
    \centering
    \resizebox{\columnwidth}{!}{%
    \begin{tabular}{lrrr}
        \toprule
         Method & AUROC$\uparrow$ & AUPRC$\uparrow$ & FPR80$\downarrow$ \\
         \midrule
         \multicolumn{4}{c}{\textbf{FashionMNIST (in) / MNIST (out)}} \\
         \midrule
         \multicolumn{4}{l}{\textbf{Use prior knowledge of OOD}} \\
Backgr. contrast. LR (PixelCNN) {[1]}               & $0.994$ & $0.993$ & $0.001$ \\
Backgr. contrast. LR (VAE) {[7]}                    & $0.924$ & - & - \\
Binary classiﬁer {[1]}                              & $0.455$ & $0.505$ & $0.886$ \\ %
$p(\hat{y} | \x)$ with OOD as noise class {[1]}     & $0.877$ & $0.871$ & $0.195$ \\ %
$p(\hat{y} | \x)$ with calibration on OOD {[1]}     & $0.904$ & $0.895$ & $0.139$ \\ %
Input complexity ($S$, Glow) [9]                    & $0.998$ & - & - \\
Input complexity ($S$, PixelCNN++) [9]              & $0.967$ & - & - \\
         \multicolumn{4}{l}{\textbf{Use in-distribution data labels $y$}} \\
$p(\hat{y} | \x)$ {[1], [2]}                        & $0.734$ & $0.702$ & $0.506$ \\
Entropy of $p(y | \x)$ {[1]}                        & $0.746$ & $0.726$ & $0.448$ \\
ODIN {[1, 3]}                                       & $0.752$ & $0.763$ & $0.432$ \\
VIB [4, 7]                                          & $0.941$ & - & - \\
Mahalanobis distance, CNN {[1]}                     & $0.942$ & $0.928$ & $0.088$ \\
Mahalanobis distance, DenseNet {[5]}                & $0.986$ & - & - \\
Ensemble, 20 classiﬁers {[1, 6]}                  & $0.857$ & $0.849$ & $0.240$ \\
         \multicolumn{4}{l}{\textbf{No OOD-specific assumptions}} \\
         \multicolumn{4}{l}{\textit{- Ensembles}} \\
WAIC, 5 models, VAE {[7]}                          & $0.766$ & - & - \\
WAIC, 5 models, PixelCNN {[1]}                      & $0.221$ & $0.401$ & $0.911$ \\
        \multicolumn{4}{l}{\textit{- Not ensembles}} \\
Likelihood regret [8]                               & $\mathbf{0.988}$ & - & - \\
$\mathcal{L}^{>0}$ + HVAE (ours)                    & $0.268$ & $0.363$ & $0.882$ \\
$\mathcal{L}^{>1}$ + HVAE (ours)                    & $0.593$ & $0.591$ & $0.658$ \\
$\mathcal{L}^{>2}$ + HVAE (ours)                    & $0.712$ & $0.750$ & $0.548$ \\
$LLR^{>1}$ + HVAE (ours)                            & $0.964$ & $0.961$ & $0.036$ \\
$LLR^{>1}_{250}$ + HVAE (ours)                      & $0.984$ & $\mathbf{0.984}$ & $\mathbf{0.013}$ \\
         \midrule
         \multicolumn{4}{c}{\textbf{CIFAR10 (in) / SVHN (out)}} \\
         \midrule
         \multicolumn{4}{l}{\textbf{Use prior knowledge of OOD}} \\
Backgr. contrast. LR (PixelCNN) {[1]}               & $0.930$ & $0.881$ & $0.066$ \\
Backgr. contrast. LR (VAE) {[8]}                    & $0.265$ & - & - \\
Outlier exposure {[9]}                              & $0.984$ & - & - \\
Input complexity ($S$, Glow) [10]                   & $0.950$ & - & - \\
Input complexity ($S$, PixelCNN++) [10]             & $0.929$ & - & - \\
Input complexity ($S$, HVAE) (Ours) [10]\footnotemark & $0.833$ & $0.855$ & $0.344$ \\
         \multicolumn{4}{l}{\textbf{Use in-distribution data labels $y$}} \\
Mahalanobis distance {[5]}                          & $0.991$ & - & -  \\
         \multicolumn{4}{l}{\textbf{No OOD-specific assumptions}} \\
         \multicolumn{4}{l}{\textit{- Ensembles}} \\
WAIC, 5 models, Glow {[7]}                          & $1.000$ & - & - \\
WAIC, 5 models, PixelCNN {[1]}                      & $0.628$ & $0.616$ & $0.657$ \\
         \multicolumn{4}{l}{\textit{- Not ensembles}} \\
Likelihood regret [8]                               & $0.875$ & - & - \\
$LLR^{>2}$ + HVAE (ours)                            & $0.811$ & $0.837$ & $0.394$ \\
$LLR^{>2}$ + BIVA (ours)                            & $\mathbf{0.891}$ & $\mathbf{0.875}$ & $\mathbf{0.172}$ \\
         \bottomrule
    \end{tabular}
    }
    \caption{
    AUROC$\uparrow$, AUPRC$\uparrow$ and FPR80$\downarrow$ for OOD detection for a FashionMNIST model using scores on the FashionMNIST test set as reference. We bold the best results within the "No OOD-specific assumptions" group since we only compare directly to those.
    HVAE (ours) refers to our hierarchical bottom-up VAE.
    BIVA (ours) refers to our implementation of the hierarchical BIVA model \cite{maaloe_biva_2019}.
    {[1]} is \cite{ren_likelihood_2019}, 
    {[2]} is \cite{hendrycks_baseline_2017}, 
    {[3]} is \cite{liang_enhancing_2018}, 
    {[4]} is \cite{alemi_uncertainty_2018}, 
    {[5]} is \cite{lee_simple_2018}, 
    {[6]} is \cite{lakshminarayanan_simple_2017}, 
    {[7]} is \cite{choi_waic_2019}, 
    {[8]} is \cite{xiao_likelihood_2020}, 
    {[9]} is \cite{hendrycks_deep_2019}, 
    {[10]} is \cite{serra_input_2020}.
    }
    \label{tab:rocauc-ood}
\end{table}

\subsection{Likelihood-based OOD detection}
\begin{figure*}[!ht]
    \begin{subfigure}[l]{0.685\columnwidth}
        \centering
        \includegraphics[width=1\columnwidth]{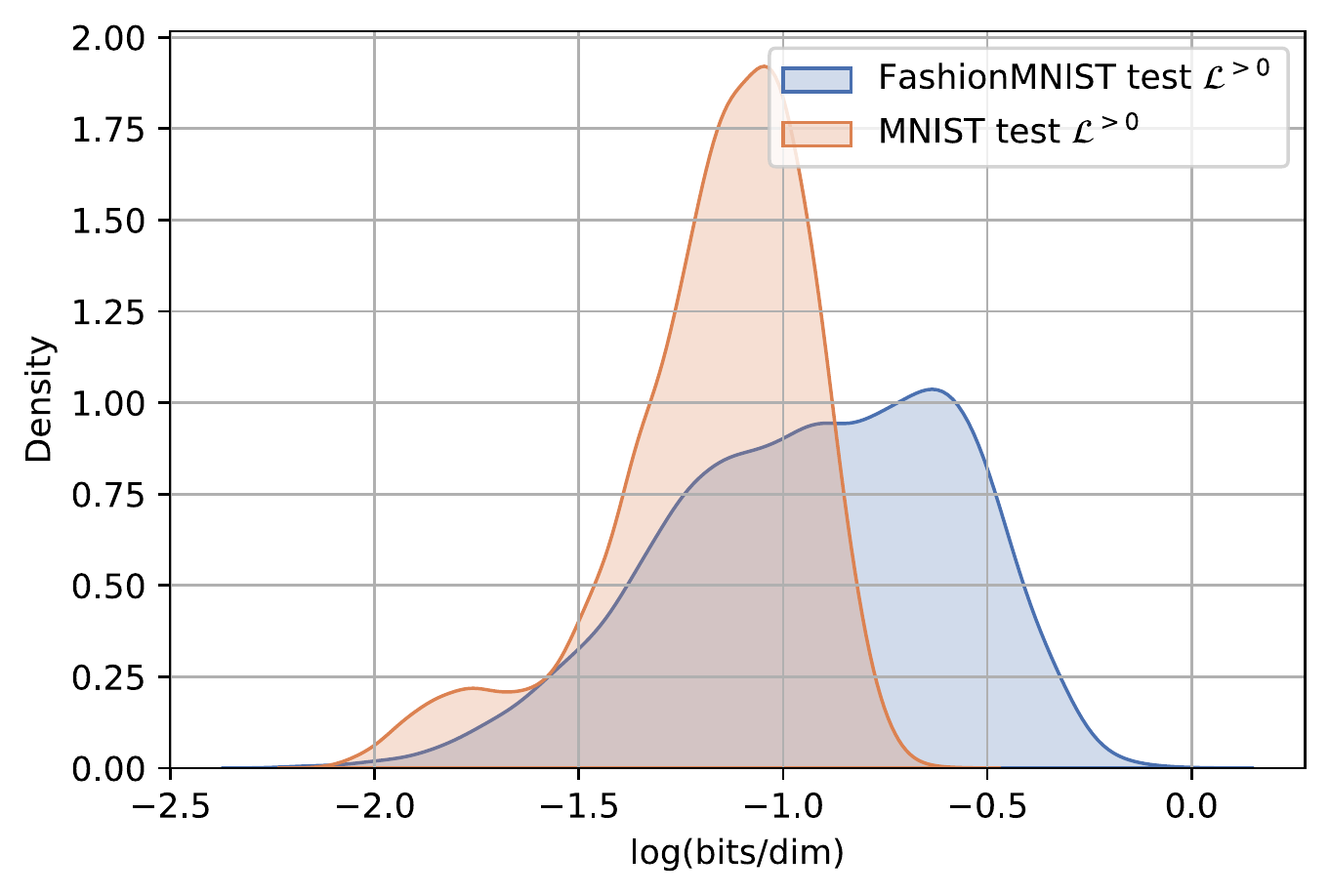}
        \caption{}
        \label{fig:FMNIST-elbo-k0}
    \end{subfigure}
    \hfill
    \begin{subfigure}[c]{0.685\columnwidth}
        \centering
        \includegraphics[width=1\columnwidth]{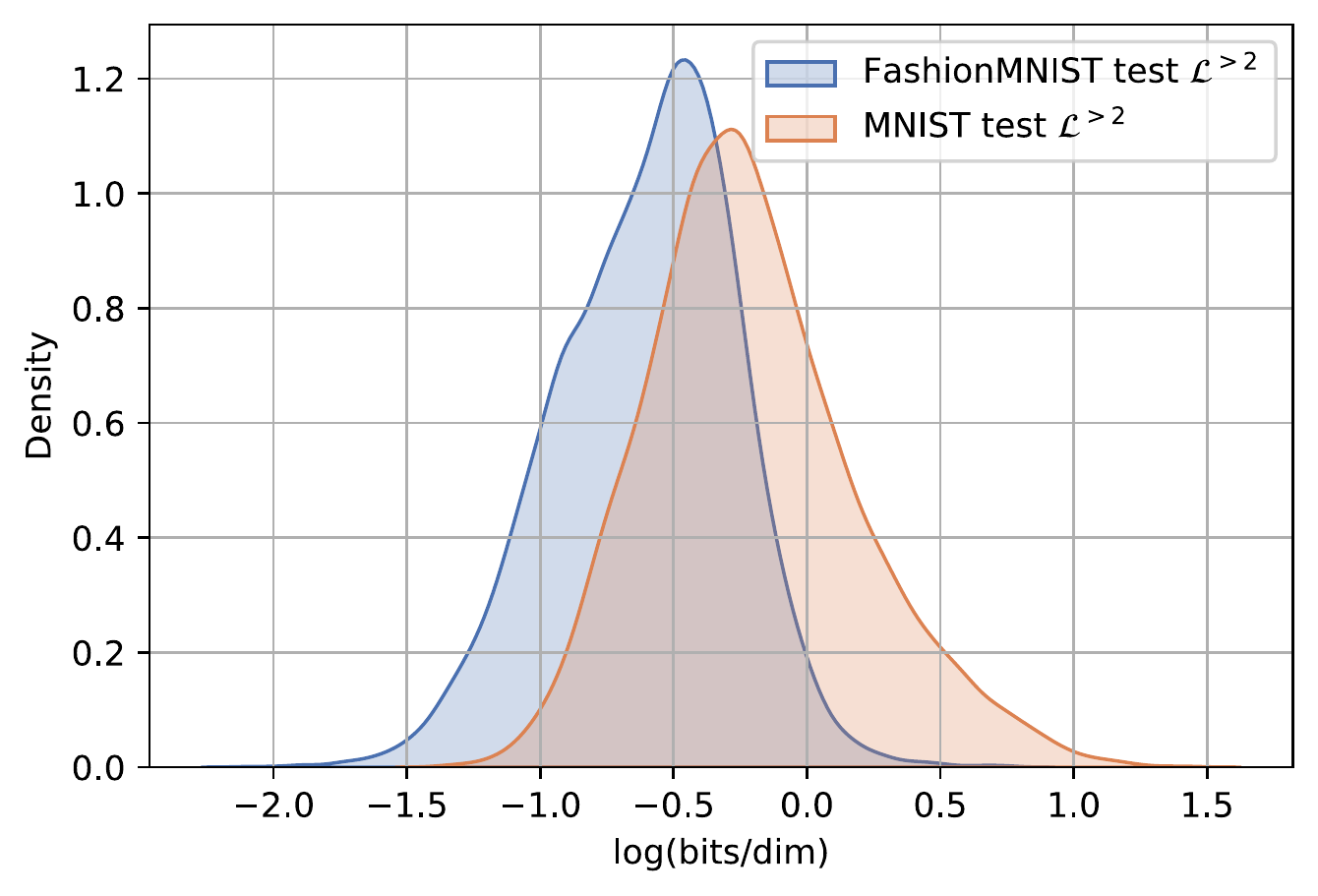}
        \caption{}
        \label{fig:FMNIST-elbo-k2}
    \end{subfigure}
    \hfill
    \begin{subfigure}[r]{0.685\columnwidth}
        \centering
        \includegraphics[width=1\columnwidth]{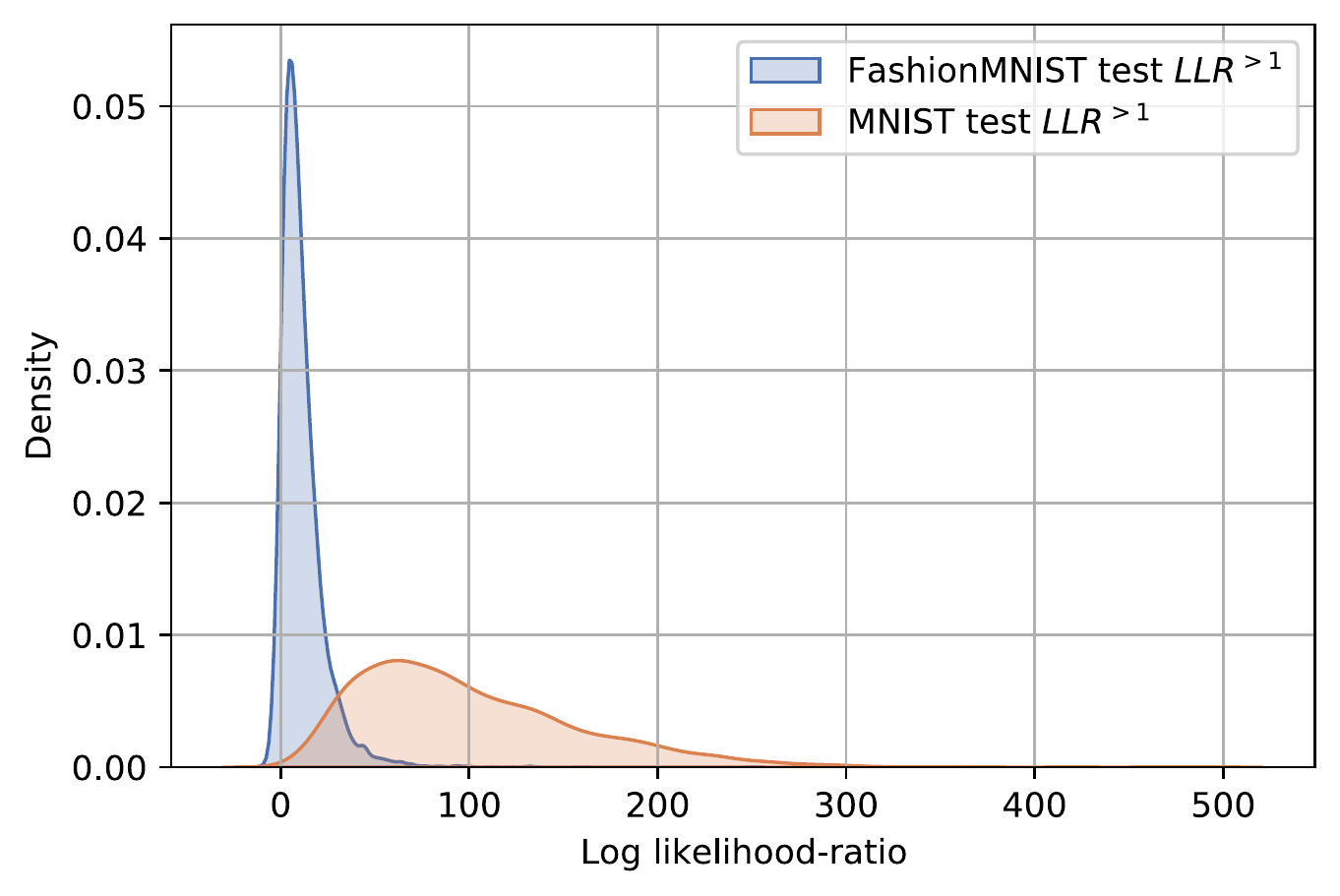}
        \caption{}
        \label{fig:FMNIST-llr}
    \end{subfigure}
    \caption{
    Empirical densities of FashionMNIST (in-distribution) and MNIST (OOD) using the raw likelihood (\subref{fig:FMNIST-elbo-k0}), the $\mathcal{L}^{>2}$ bound (\subref{fig:FMNIST-elbo-k2}) and the $LLR^{>1}$ score (\subref{fig:FMNIST-llr}). All densities are computed using the HVAE model.).
    For the regular likelihood MNIST is very clearly more likely on average than the FashionMNIST test data while with the $\mathcal{L}^{>2}$ bound separation is better but significant overlap remains.
    The $LLR^{>1}$ provides a high degree of separation. Likelihoods are reported in units of the natural log of the number of bits per dimension.
    }
    \label{fig:FMNIST-ood-densities}
\end{figure*}

We first report the results of the different variations of the $\mathcal{L}^{>k}$ bound for OOD detection. 
We reconfirm the results of \citet{nalisnick_deep_2019} by observing that our hierarchical latent variable models also assign higher $\mathcal{L}^{>0}$ to the OOD dataset in the FashionMNIST/MNIST and CIFAR10/SVHN cases resulting in an AUROC$\uparrow$ inferior to random (\autoref{tab:rocauc-ood}).
Switching the in-distribution data for the OOD data in both cases result in correctly detecting the OOD data; an asymmetry also reported by \citet{nalisnick_deep_2019}.
\autoref{fig:FMNIST-elbo-k0} shows the density of $\mathcal{L}^{>0}$ in bits per dimension \cite{theis_note_2016} by the model trained on FashionMNIST when evaluated on the FashionMNIST and MNIST test sets.
We observe a high degree of overlap, with less separation of the OOD data compared to similar results of autoregressive and flow-based models, like \citet{xiao_likelihood_2020}.

We then evaluate the looser $\mathcal{L}^{>k}$ \eqref{eq:biva->k} for $k\in\{1,L\}$.
\autoref{fig:FMNIST-elbo-k2} shows the result for $\mathcal{L}^{>2}$, which yielded the highest AUCROC$\uparrow$, only slightly better than random.
Like \citet{maaloe_biva_2019}, we see that increasing the value of $k$ generally leads to improved OOD detection.
However, we also observe that the two empirical distributions never cease to overlap.
Importantly, depending on the OOD dataset, the amount of remaining overlap can be high which limits the discriminatory power of the likelihood-based $\mathcal{L}^{>k}$ bound.
This is in-line with the pathological behavior of the raw likelihood of latent variable models when used for OOD detection \cite{xiao_likelihood_2020}.
Since a high degree of overlap also seems present in \citet{maaloe_biva_2019}, and we see the same problem for our BIVA model trained on CIFAR10, we do not expect this to be due to the less expressive HVAE.

\subsection{Likelihood-ratio-based OOD detection}
\begin{figure}
    \centering
    \includegraphics[width=\columnwidth]{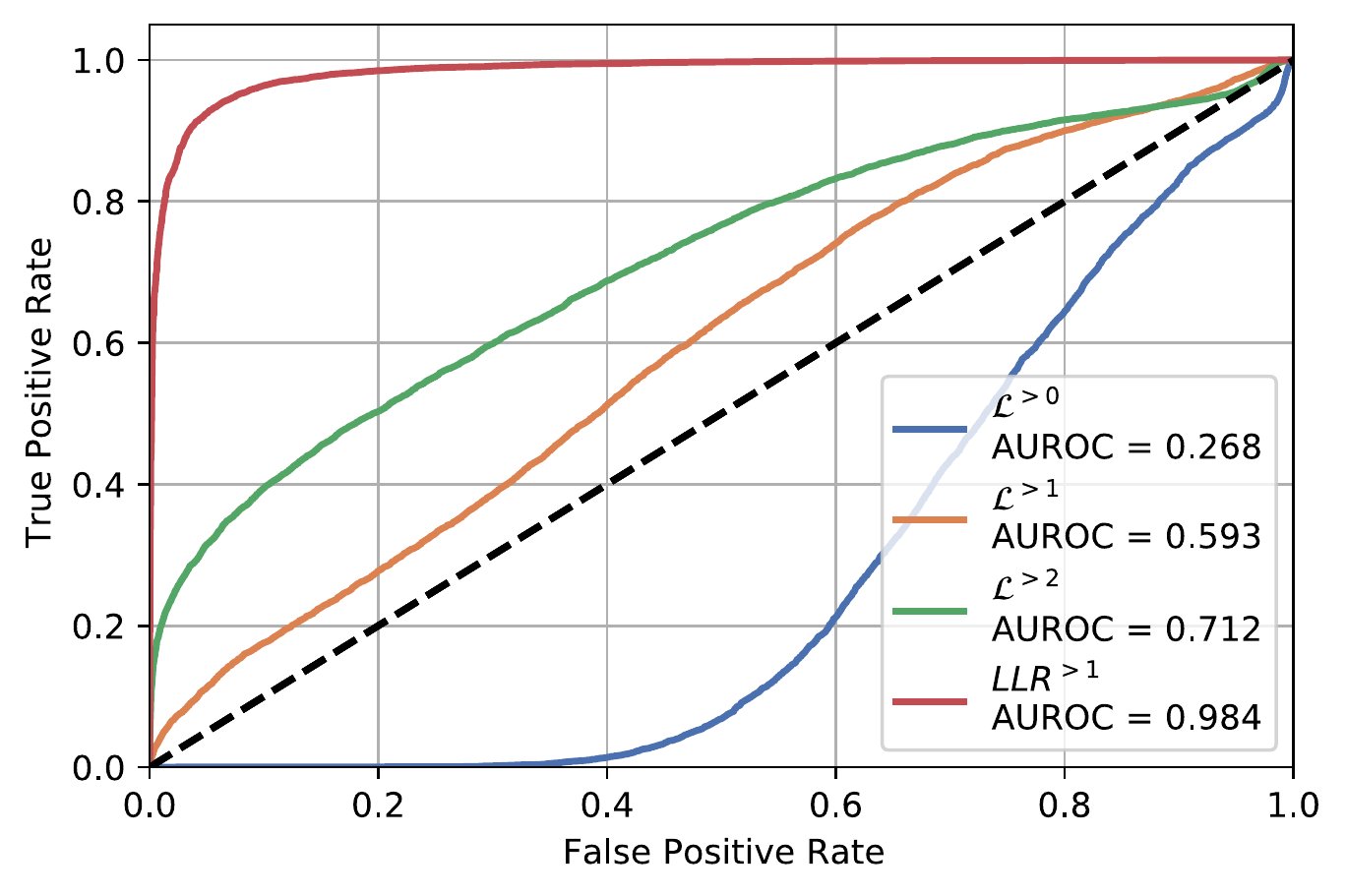}
    \caption{ROC curves with AUROC score for detecting MNIST as OOD with the HVAE model trained on FashionMNIST.
    A ROC curve is plotted for each of the $\mathcal{L}^{>k}$ bounds including the ELBO along with one for the best-performing log likelihood-ratio $LLR^{>1}$.}
    \label{fig:FMNIST-roc-llr}
\end{figure}
\begin{figure}
    \centering
    \includegraphics[width=\columnwidth]{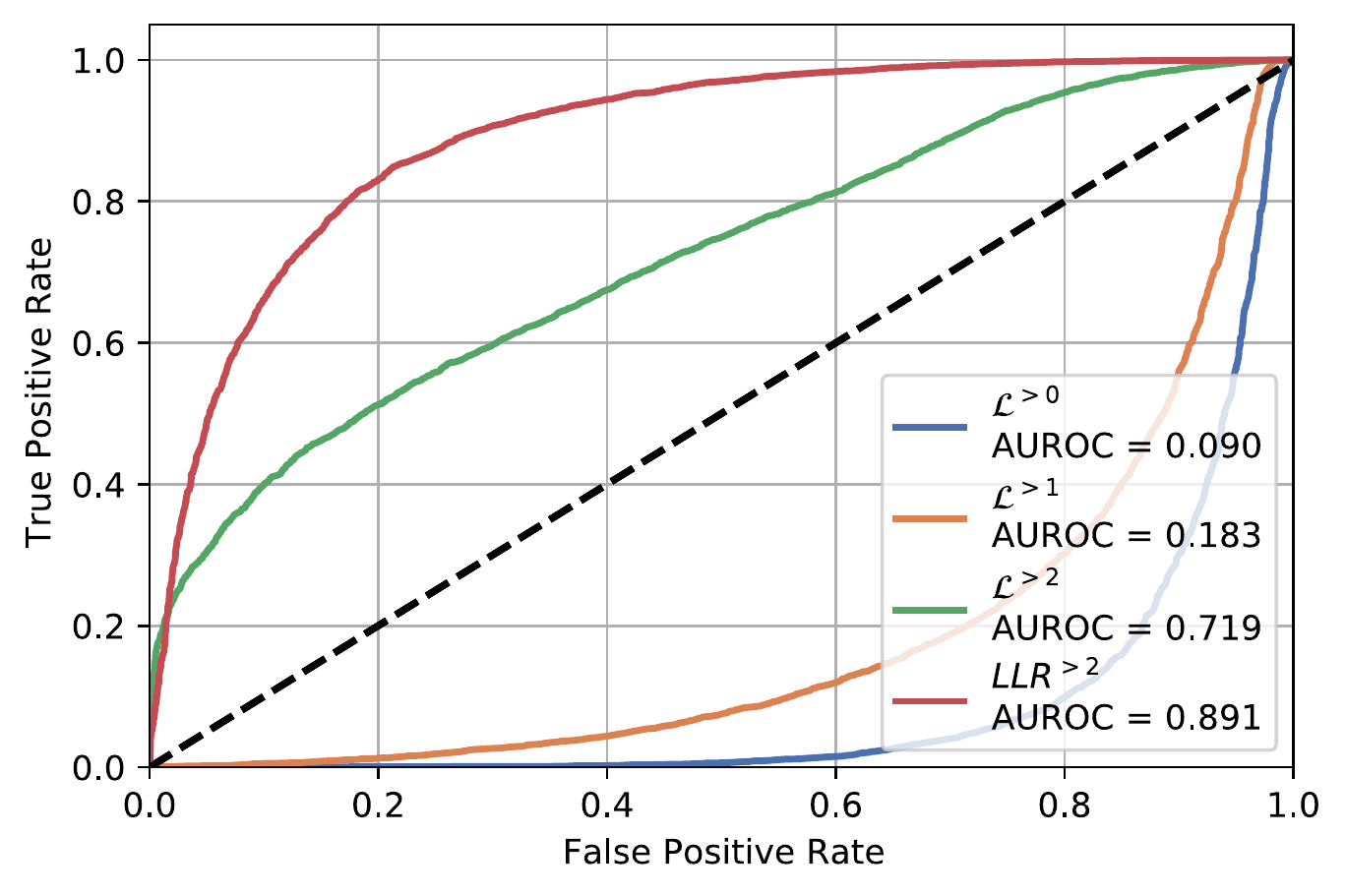}
    \caption{ROC curves with AUROC score for detecting SVHN as OOD with the BIVA model trained on CIFAR10.
    A ROC curve is plotted for each of the $\mathcal{L}^{>k}$ bounds including the ELBO along with one for the best-performing log likelihood-ratio $LLR^{>2}$.}
    \label{fig:CIFAR10-roc-llr}
\end{figure}
We now move to the likelihood ratio-based score.
We find that $LLR^{>k}$ separates the OOD MNIST data from in-distribution FashionMNIST to a higher degree than the likelihood estimates as can be seen by the empirical densities of the score in \autoref{fig:FMNIST-llr}.
We note that the likelihood ratio between the ELBO and the $\mathcal{L}^{>k}$ bound provides the highest degree of separation of MNIST and FashionMNIST as measured by the AUROC$\uparrow$ for $k=1$ smaller than $L$.
This is not surprising since the value of $k$ that provides the maximal separation to the reference in-distribution dataset need not be the one for which $\mathcal{LLR}^{>k}$ is overall maximal for the OOD dataset.
We also visualize the ROC curves resulting from using the $LLR^{>k}$ score for OOD detection on both FashionMNIST/MNIST and CIFAR10/SVHN and compare it to the ROC curves resulting from the different $\mathcal{L}^{>k}$ bounds in Figures \ref{fig:FMNIST-roc-llr} and \ref{fig:CIFAR10-roc-llr}, respectively.
On both datasets we see significantly better discriminatory performance when using the $LLR^{>k}$ score.

\autoref{tab:rocauc-ood} shows that BIVA improves upon the HVAE model for OOD detection on CIFAR while \autoref{tab:bits-per-dim-ood} shows that the BIVA model also improves upon the HVAE in terms of likelihood.
We hypothesize that models larger than our implementation of BIVA, with better likelihood scores may perform even better \cite{maaloe_biva_2019, vahdat_nvae_2020, child_very_2021}.

\subsection{Comparison to baselines}
\textbf{Performance}: 
\autoref{tab:rocauc-ood} summarize our results compared to baselines based on the commonly used AUROC$\uparrow$, AUPRC$\uparrow$ and FPR80$\downarrow$ metrics.
Our method outperforms other generative model-based methods such as WAIC \cite{choi_waic_2019} with Glow model and performs similarly to the likelihood regret method of \cite{xiao_likelihood_2020}.
Furthermore, our method performs similarly to the background constrative likelihood ratio method of \citet{ren_likelihood_2019} on FashionMNIST/MNIST but contrary to the failure of that method on CIFAR10/SVHN reported by \cite{xiao_likelihood_2020}, our method performs very well on this task too.
Our approach outperforms all supervised approaches that use in-distribution labels or synthetic examples of OOD data derived from the in-distribution data including ODIN \cite{liang_enhancing_2018} and the predictive distribution of a classifier $p(\hat{y}|\x)$ trained and evaluated in various ways (see \citet{ren_likelihood_2019}).

\textbf{Runtime}:
For a full evaluation of a single example across all feature levels of a model with $L$ stochastic layers, our method requires $L-1$ forward passes through the inference and generative networks as well as computing the likelihood ratio, of which the forward passes are dominant.
For a typical forward pass that is linear in the input dimensionality, $D$, and the number of stochastic layers, $L$, this amounts to computation of $\mathcal{O}(DL)$.
Compared to some related work that either requires an $M>1$ sized batch of inputs of which either all or none are OOD \cite{nalisnick_detecting_2019} or cannot be applied to batches due to the required per-example optimization \cite{xiao_likelihood_2020}, our method additionally is applicable to batches of any size that may consist of both OOD and in-distribution examples which provides drastic speed-ups via vectorization and parallelization.
Furthermore, the method of \citet{xiao_fashion-mnist_2017} requires refitting the inference network of a VAE which can be computationally demanding.
Compared to the likelihood ratio proposed in \citet{ren_likelihood_2019}, our method requires training only a single model on a single dataset.

\section{Discussion}
Deep generative models are state-of-the-art density estimators, but the OOD failures reported in recent years have raised concerns about the limitations of such density estimates. Recent work on improving OOD detection has largely sidestepped this concern by relying on additional assumptions that strictly should not be needed for models with explicit likelihoods.
While the engineering challenge of building reliable OOD detection schemes is important, it is of more fundamental importance to understand \emph{why} the naive likelihood test fails.
We have provided evidence that low-level features of the neural nets dominate the likelihood, which gives a \emph{cause} to the \emph{why}.
The fact that a simple score for measuring the importance of semantic features yield state-of-the-art results on OOD detection without access to additional information gives validity to our hypothesis.

The findings from, amongst others, \citet{nalisnick_deep_2019, serra_input_2020} have a clear relation to information theory and compression. 
Semantically complex in-distribution data yields models with diverse low-level feature sets that enable generalization across datasets.
Simpler datasets can only yield models with less diverse low-level feature sets compared to complex training data.
Hence, there can be an asymmetry where the likelihoods of simple OOD data can be high for a model trained on complex data, but not the other way around.
Loosely put, the minimal number of bits required to losslessly compress data sampled from some distribution is the entropy of the generating process \cite{shannon_mathematical_1948, mackay_information_2003}.
\citet{townsend_practical_2019} recently showed that VAEs can be used for lossless compression at rates superior to more generic algorithms.

We also note that since the hierarchical VAE is a probabilistic graphical latent variable model, it lends itself very naturally to manipulation at the feature level \citep{kingma_semi-supervised_2014, maaloe_auxiliary_2016, maaloe_semi-supervised_2017}.
This property sets it apart from other generative models that do not explicitly define such a hierarchy of features.
This in turn enables reliable OOD detection with our methodology while making no explicit assumptions about the nature of OOD data and only using a single model. This has not been achieved with autoregressive or flow-based models.

\section{Conclusion}
In this paper we study unsupervised out-of-distribution detection using hierarchical variational autoencoders.
We provide evidence that highly generalizeable low-level features contribute greatly to estimated likelihoods resulting in poor OOD detection performance.
We proceed to develop a likelihood-ratio based score for OOD detection and define it to explicitly ensure that data must be in-distribution across all feature levels to be regarded in-distribution.
This ratio is mathematically shown to perform OOD detection in the latent space of the model, removing the reliance on the troublesome input-space likelihood.
We point out that contrary to much recent literature on OOD detection, our approach is fully unsupervised and does not make assumptions about the nature of OOD data.
Finally, we demonstrate state-of-the-art performance on a wide range of OOD failure cases.

\section*{Acknowledgements}
This research was partially funded by the Innovation Fund Denmark via the Industrial PhD Programme (grant no.\@ 0153-00167B). JF and SH were funded in part by the Novo Nordisk Foundation (grant no.\@ NNF20OC0062606) via the Center for Basic Machine Learning Research in Life Science (MLLS, \hyperlink{https://www.mlls.dk}{https://www.mlls.dk}). JF was further funded by the Novo Nordisk Foundation (grant no.\@ NNF20OC0065611) and the Independent Research Fund Denmark (grant no.\@ 9131-00082B). SH was further funded by VILLUM FONDEN (15334) and the European Research Council (ERC) under the European Union’s Horizon 2020 research and innovation programme (grant agreement no. 757360).

\bibliography{references.bib}
\bibliographystyle{icml2021}

\clearpage
\appendix
\section{Datasets}
\autoref{tab:datasets-overview} lists the datasets used in the paper. We use the predefined train/test splits for the datasets.

For SmallNORB and Omniglot we resize the original grey-scale images to $28\times28$ with ordinary bi-linear interpolation.
For each of these datasets, we also create a version where the grey-scale is inverted. We do this because, the overall white nature of the images tends to make detecting them as OOD from FashionMNIST artificially easy.
The inversion is done via the simple transformation $\x_\text{inverted} = 255 - \x_\text{original}$ since images are encoded as 8 bit unsigned integers.

\begin{table}[h!]
    \centering
    \resizebox{\columnwidth}{!}{%
    \begin{tabular}{llr}
        \toprule
        Dataset & Dimensionality & Examples \\
        \midrule
        FashionMNIST \cite{xiao_fashion-mnist_2017} & $28\times28\times1$ & 70,000 \\
        MNIST \cite{lecun_gradient-based_1998} & $28\times28\times1$ & 70,000 \\
        notMNIST \cite{bulatov_notmnist_2011} & $28\times28\times1$ & 547,838 \\
        KMNIST \cite{clanuwat_deep_2018} & $28\times28\times1$ & 70,000 \\
        Omniglot \cite{lake_human-level_2015} & $28\times28\times1$ & 32,460 \\
        SmallNORB \cite{lecun_learning_2004} & $28\times28\times1$ & 97,200 \\
        \midrule
        CIFAR10 \cite{krizhevsky_learning_2009} & $32\times32\times3$ & 60,000 \\
        SVHN \cite{netzer_reading_2011} & $32\times32\times3$ & 99,289 \\
        \bottomrule
    \end{tabular}
    }
    \caption{Overview of the used datasets.}
    \label{tab:datasets-overview}
\end{table}

\section{Model details}\label{sec:model-details}
In \autoref{tab:hyperparameters} we specify the hyperparameters used when training our models.

We make our source code available at \url{https://github.com/JakobHavtorn/hvae-oodd}.

\subsection{Hierarchical VAE}
Our Hierarchical VAE (HVAE) model uses bottom-up inference and top-down generative paths as specified in the paper.
For grey-scale images, the output is parameterized by a Bernoulli distribution while for natural images we use a Discretized Logistic Mixture \cite{salimans_pixelcnn_2017}.
The latent variables are parameterized by stochastic layers that output the mean and log-variance of a diagonal covariance Gaussian. The prior distribution on the top-most latent is a standard Gaussian.
For grey-scale images, the lowest latent space is parameterized by a convolutional neural network and has dimensions $14\times14\times8$ interpreted as (height $\times$ width $\times$ latent dimension). The highest two latent variables are parameterized by dense transformations with $16$ and $8$ units, respectively.
For natural images, the bottom-two latent variables are parameterized by convolutional neural networks and have dimensions $(16\times16)\times128$, $(8\times8)\times64$, respectively for $\z_1, \z_2$. The top-most latent, $\z_3$, is densely connected with dimension $32$.

Each stochastic layer is preceeded by a determininistic transformation.
For both grey-scale and natural images, each deterministic transformation consists of three residual blocks of the same type used by \citet{maaloe_biva_2019}. The structure of a residual block is:
\begin{equation*}
    \y = \text{Conv}\left( \text{Act} \left( \text{Conv}_s \left( \text{Act}(\x) \right) \right) \right) + \x \ ,
\end{equation*}
where ``Conv'' refers to a same-padded convolution and ``Act'' to the activation function. Within a residual block, the first convolution always has stride 1 while the second convolution has stride $s$. In a deterministic transformation, any non-unit stride is performed in the third residual block. For grey-scale images, we stride by 2 in the first and second deterministic transformations but not the third. For natural images, we similarly stride by 2 in the first and second deterministic transformations. For grey-scale we use 64 channels while we use 256 for natural images.
In both cases, the first deterministic block uses a kernel size of 5 and the latter two a kernel of size 3. We use the ReLU activation function \cite{fukushima_neocognitron_1980, nair_rectified_2010}.

Since the benefits and drawbacks of using batch normalization \cite{ioffe_batch_2015} in hierarchical VAEs is still the matter of some debate \cite{sonderby_ladder_2016, vahdat_nvae_2020, child_very_2021} we choose to use weight normalization \cite{salimans_weight_2016} as in other work \cite{maaloe_biva_2019} and initialize the model using the originally proposed data-dependent initialization.
To have the stochastic layers initialize to standard Gaussian distributions (zero mean, unit variance), with this initialization, we select the activation function for the variance as a Softplus,
\begin{equation*}
    \text{Softplus}(\x) = \frac{1}{\beta}\log\left(1 + \exp(\beta\x)\right) \ ,
\end{equation*}
with $\beta = \log(2) \approx 0.693$ to output 1 for $\x = 0$.

Training of a HVAE model took approximately two days on a single NVIDIA GTX 1080 Ti graphics card.

\subsection{BIVA}
For the BIVA model \cite{maaloe_biva_2019}, we use a specification that is very similar to that of the HVAE above, and to that of the original paper.
The model has 10 latent variables the lowest 3 of which are spatial and the rest are densely connected in order to have an architecture similar to the HVAE.
The model uses an overall stride of 8, achieved by striding by 2 in the first, fourth and sixth deterministic transformations.
From $\z_1$ to $\z_{10}$, the latents have the following dimensions: The lowest three latents are spatial $(16\times16)\times8$, $(16\times16)\times16$ and $(16\times16)\times32$, given as $(\text{height}\times\text{width})\times\text{dim})$, while the rest are dense vectors with dimensions of $42, 40, 38, 36, 34, 32, 30$.

Training of a BIVA model took approximately a week on a single NVIDIA GTX 1080 Ti graphics card.

\begin{table}[t]
    \centering
    \resizebox{\columnwidth}{!}{%
    \begin{tabular}{ll}
        \toprule
        Hyperparameter & Setting/Range \\
        \midrule
        \multicolumn{2}{l}{\textbf{All}} \\
        \midrule
        Optimization & Adam \cite{kingma_adam_2015} \\
        Learning rate & $3e-4$ \\
        Batch size & 128 \\
        Epochs & 2000 \\
        Free bits & $\SI{2}{nats}$ per $\z_i$ shared across latent dim. \\
        Free bits constant & 200 epochs \\
        Free bits annealed & 200 epochs \\
        Activation & ReLU \\
        \multirow{2}{*}{Initialization} & Data-dependent \\
                                        & \cite{salimans_weight_2016} \\
        \midrule
        \multicolumn{2}{l}{\textbf{HVAE}} \\
        Latent dimensionality & 128-64-32 (natural) / 8-16-8 (grey) \\
        Convolution kernel & 5-3-3 \\
        Stride & 2-2-1 \\
        Warmup anneal period & 200 epochs \\
        \midrule
        \multicolumn{2}{l}{\textbf{BIVA}} \\
        \multirow{2}{*}{Latent dimensionality} & 10-8-6 (spatial) \\
                                               & 42-40-38-36-34-32-30 (dense) \\
        Convolution kernel & 5-3-3-3-3-3-3-3-3-3 \\
        Stride & {2-1-1-2-1-2-1-1-1-1} \\
        \bottomrule
    \end{tabular}
    }
    \caption{Selection of most important hyperparameters and their setting. Convolutional kernels are square and latent dimensions are given without spatial dimensions which are given in the text. See \autoref{sec:model-details} for more details.}
    \label{tab:hyperparameters}
\end{table}

\section{Analysis of the influence of latent variables on the marginal likelihood}
In the paper, we argue that the lowest level latent variables, which have the highest dimensionality, contribute the most to the approximate likelihood.
Here, we provide a stringent mathematical argument that generalizes this to the exact marginal likelihood in a model with a deterministic decoder.

\subsection{Model specification}
For an arbitrary hierarchical latent variable model, we have a prior $p(\z_L)$ and a generative mapping $f: \mathbb{R}^d \rightarrow \mathbb{R}^D$, such that $\x = f(\z_L)$ and $D>d$.
Note that we will assume that $f$ is deterministic, such that we are effectively working with $p(\x|\z) = \delta_{f(\z)}(\x)$.
This is a limiting assumption, but it allows working through the following. For shorthand we will simply write $\z=\z_L$.

Let $f$ have a bottleneck architecture, i.e.\
\begin{align}
f(\z) &= f_1(\ldots f_{L-1}(f_L(\z))) \ ,
\end{align}
where
\begin{align}
f_i: \mathbb{R}^{d_{i}} \rightarrow \mathbb{R}^{d_{i-1}}, \qquad i = L, \ldots, 1 \ .
\end{align}
Here we use the notation $d_0=D=\vert\x\vert$ and $d_L=d=\vert\z\vert$ and further assume $d_0\geq d_1\geq \ldots \geq d_{L-1} \geq d_L$ which gives the bottleneck.

Assuming $\x$ is such that a corresponding latent variable $\z$ exists, i.e.\ that there exists $\z$ such that $\x = f(\z)$, then we can write the likelihood of $\x$ through a standard change of variables (similar to flow-based models),
\begin{align}
p(\x) &= p(\z) \prod_{i = 1}^L \left(\sqrt{\det \J_i^T \J_i}\right)^{-1} \ ,
\end{align}
where $\J_i$ is the Jacobian of $f_i$, i.e.
\begin{align}
\J_i &= \frac{\partial f_i}{\partial \z_i} \in \mathbb{R}^{d_i\times d_{i-1}} \ .
\end{align}
Here we use the notation that $\z_i$ is the representation at layer $i$.
Note that $\J_i^T \J_i$ is a $d_{i-1} \times d_{i-1}$ symmetric positive semidefinite matrix (determinant $\geq 0$).

The log-likelihood can be written as
\begin{equation}
    \log p(\x) = \log p(\z) - \frac{1}{2} \sum_{i=1}^L \log \det \J_i^T\J_i \ .
\end{equation}

By construction of determinants, we can generally expect these determinants to grow with the dimensionality of the matrix.
We should expect the determinant of a $d \times d$ matrix to be of the order $\mathcal{O}(\lambda^d)$ for some number $\lambda>0$.
With that in mind, we should generally expect that
\begin{align}
\det \J_{i+1}^T \J_{i+1} < \det \J_i^T \J_i \ ,
\end{align}
due to the bottleneck assumption.
If so, we see that the marginal likelihood $p(\x)$ will be dominated by $\left(\sqrt{\det \J_1^T \J_1}\right)^{-1}$, i.e.\ low-level features have a higher influence on the likelihood than more important semantic ones.

\subsection{The Gaussian case}
\begin{figure}[t]
  \centering
  \includegraphics[width=\columnwidth]{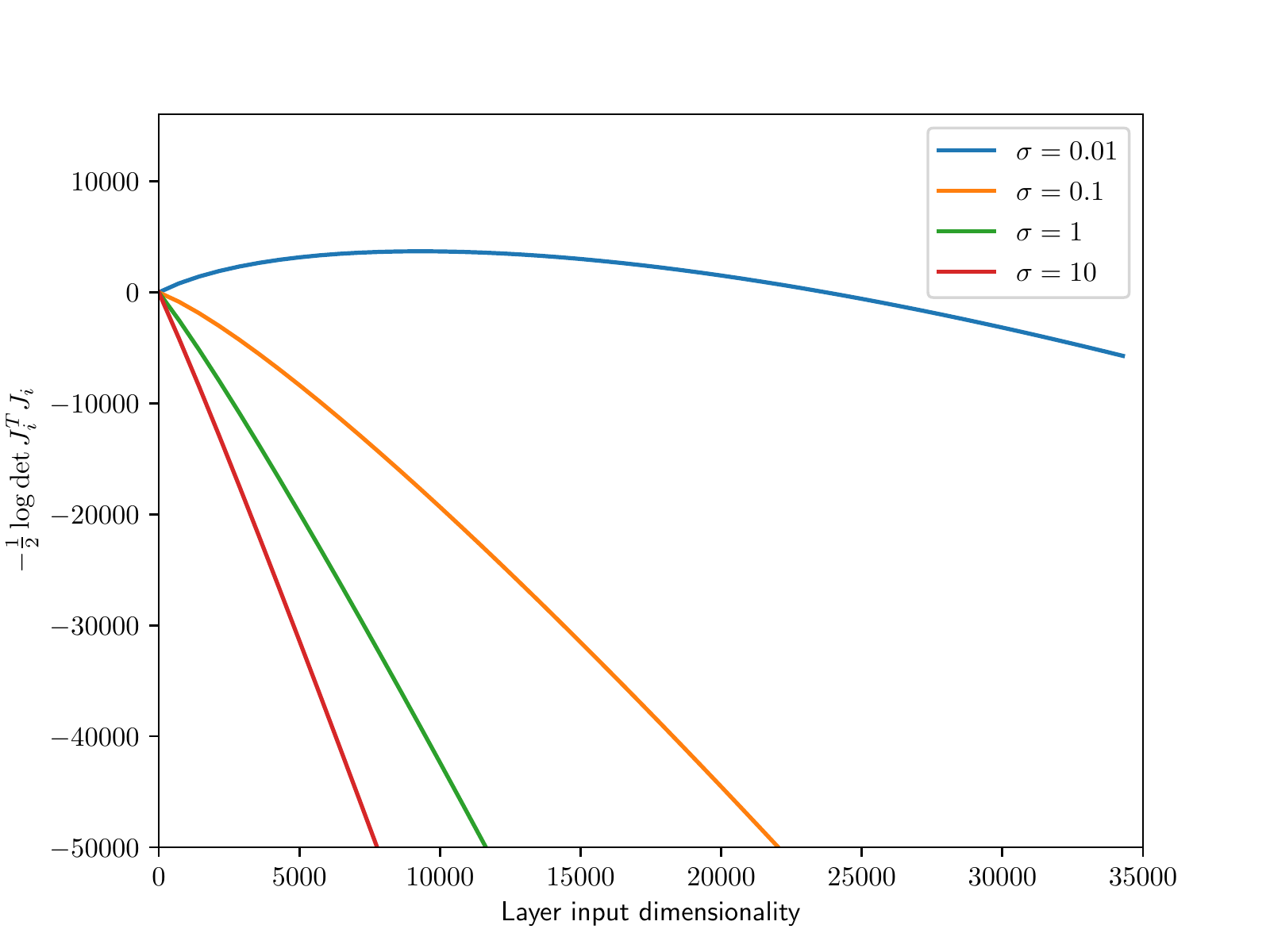}
  \caption{The expected inverse volume change for Gaussian Jacobians \eqref{eq:Edet} on a log-scale.}
  \label{fig:Edet}
\end{figure}
The previous remarks can be made more precise if we make distributional assumptions on the Jacobians.
Here we will assume that the Jacobians of each layer follow a Gaussian distribution.
Specifically, we will assume that each entry in $\J_i$ is distributed as $\mathcal{N}(0, \sigma^2)$.
The analysis below extends to nonzero means and more general covariance structure, but this comes with a cost of less transparent notation.
In this setting, $\J_i^T \J_i$ follows a Wishart distribution (in the general setting it would follow a non-central Wishart distribution).
\citet{muirhead_aspects_2009} tells us that the expected multiplicative contribution to the likelihood of each layer is
\begin{align}
\mathbb{E}\left[ \left(\sqrt{\det \J_i^T \J_i}\right)^{-1} \right]
  &= \sigma^{-d_{i-1}}  2^{-\frac{d_{i-1}}{2}} \frac{\Gamma_{d_{i-1}}\left( \frac{1}{2} d_{i} - \frac{1}{2} \right)}{\Gamma_{d_{i-1}}\left( \frac{1}{2} d_{i} \right)} \notag \\
  &= \sigma^{-d_{i-1}}  2^{-\frac{d_{i-1}}{2}} \frac{\Gamma\left( \frac{1}{2} (d_{i} - d_{i-1}) \right)}{\Gamma\left( \frac{1}{2} d_{i} \right)}
\label{eq:Edet}
\end{align}
where $\Gamma_d$ is the multivariate Gamma function.
Assuming that the increase in layer dimension $d_{i} - d_{i-1}$ is constant, then we see that \eqref{eq:Edet} goes to zero as $d_{i}$ goes to infinity as the $\Gamma$ function grows super-exponentially to infinity.
This super-exponential growth further implies that the first layers dominate the marginal likelihood $p(\x)$.
This is also visually evident in \autoref{fig:Edet}.

\section{Derivation of the $\mathcal{L}^{>k}$ bound}
In this section we present the derivation of $\mathcal{L}^{>k}$ and show that it is a lower bound on the marginal likelihood.

First, we consider a two-layered VAE with bottom-up inference.
We proceed very similarly to the derivation of the regular ELBO and also use Jensen's inequality.
\begin{align}
\log p(\x) &= \log \int\int p(\x|\z_1)p(\z_1|\z_2)p(\z_2) \text{d}\z_1\text{d}\z_2 \\
&= \log \int\int \frac{q(\z_2|\x)}{q(\z_2|\x)} p(\x|\z_1)p(\z_1|\z_2)p(\z_2) \text{d}\z_1\text{d}\z_2 \notag \\
&= \log \int\int q(\z_2|\x) p(\z_1|\z_2) \frac{p(\x|\z_1)p(\z_2)}{q(\z_2|\x)} \text{d}\z_1\text{d}\z_2 \notag \\
&\ge \mathbb{E}_{p(\z_1|\z_2)q(\z_2|\x)} \left [ \log \frac{p(\x|\z_1)p(\z_2)}{q(\z_2|\x)} \right ] \equiv \mathcal{L}^{> 1} \notag \ .
\end{align}
Here, we have introduced the variational distribution $q(\z_2|\x)$ which, naively, is different from any of the available variational distributions $q(\z_1|\x)$ and $q(\z_2|\z_1)$.
However, it's easy to see that we can simply define $q(\z_2|\x)=q(\z_2|d_1(\x))$ where $d_1(\x) = \mathbb{E}\left[q(\z_1|\x)\right]$. I.e.\@ we compute the distribution over $\z_2$ via the mode of $q(\z_1|\x)$.
This is possible since we exclusively manipulate the variational proposal distribution without altering the generative model $p(\x,\z)$.

In general, the derivation of $\mathcal{L}^{>k}$ for an $L$-layered hierarchical VAE with $\z=\z_1,\dots,\z_L$ is as follows:
\begin{align}
    \log p(\x) &= \log \int p(\x|\z)p(\z) \text{d}\z \\
    &= \log \int \frac{q(\z_{>k}|\x)}{q(\z_{>k}|\x)} p(\x|\z)p(\z) \text{d}\z \notag \\
    &= \log \int q(\z_{>k}|\x) p(\z) \frac{p(\x|\z)}{q(\z_{>k}|\x)} \text{d}\z \notag \\
    &= \log \int q(\z_{>k}|\x) p(\z_{\leq k}|\z_{>k}) p(\z_{>k}) \frac{p(\x|\z)}{q(\z_{>k}|\x)} \text{d}\z \notag \\
    &= \log \int q(\z_{>k}|\x) p(\z_{\leq k}|\z_{>k}) \frac{p(\x|\z)p(\z_{>k})}{q(\z_{>k}|\x)} \text{d}\z \notag \\
    &\ge \mathbb{E}_{p(\z_{\leq k}|\z_{>k})} \left[ \log q(\z_{>k}|\x) \frac{p(\x|\z)p(\z_{>k})}{q(\z_{>k}|\x)} \right] \notag \\
    &\ge \mathbb{E}_{p(\z_{\leq k}|\z_{>k})q(\z_{>k}|\x)} \left[ \log \frac{p(\x|\z)p(\z_{>k})}{q(\z_{>k}|\x)} \right] \equiv \mathcal{L}^{>k} \notag \ .
\end{align}
Similar to the $L=2$ case above, we have defined
$$q(\z_{>k}|\x) = q(\z_{>k}|d_k(\x))$$
with $d_k$ defined recursively as
$$d_k(\x) = \mathbb{E}\left[q(\z_k|d_{k-1}(\x))\right], \qquad d_0(\x) = \x\ .$$
That is, we simply consider the inference network below $\z_{k+1}$ to be a deterministic encoder and forward pass the mode of each preceding variational distribution.

Additionally, we obtain $p(\z_{\leq k}|\z_{>k}) p(\z_{>k})$ by splitting
$$p(\z)=p(\z_L)p(\z_{L-1}|\z_L)\cdots p(\z_{1}|\z_2)$$
at index $k$. Importantly, we then evaluate
$$p(\z_{>k})=p(\z_L)p(\z_{L-1}|\z_L)\cdots p(\z_{k+1}|\z_{k+2})$$
with samples from $q(\z_{>k}|\x)$ while
$$p(\z_{\leq k}|\z_{>k})=p(\z_{k}|\z_{k+1})p(\z_{k-1}|\z_{k})\cdots p(\z_{1}|\z_{2})$$
is evaluated for $\z_k$ with $\z_{k+1} \sim q(\z_{>k}|\x)$ and for $\z_{<k}$ with $\z_{>k}$ obtained conditionally from itself.

\section{The complementary $\mathcal{L}^{<l}$ bound}
We can generalize the $\mathcal{L}^{>k}$ bound by introducing the flipped version, $\mathcal{L}^{<l}$, which compared to $\mathcal{L}^{>k}$, instead samples the $L-l$ \textit{highest} latent variables in the hierarchy from the prior $\z_{l},\dots,\z_L \sim p_\theta(\z_{\geq l})=p_\theta(\z_l|\z_{l+1})\cdots p_\theta(\z_L)$ and the remaining lower latents from the approximate posterior $\hat{\z}_{1},\dots,\hat{\z}_{l-1} \sim q_\phi(\z_{<l}|\x)=q_\phi(\z_{1}|\x)q_\phi(\z_{2}|\z_1)\cdots q_\phi(\z_{l-1}|\z_{l-2})$,
\begin{equation}\label{eq:biva-<l}
    \mathcal{L}^{<l} = \mathbb{E}_{p_\theta(\z_{\geq l}) q_\phi(\z_{<l}|\x)} \left[ \log \frac{p_\theta(\x,\z)p_\theta(\z_{<l})}{q_\phi(\z_{<l}|\x)} \right].
\end{equation}
Similar to $\mathcal{L}^{>k}$, we recover the regular ELBO for $l=L$. Contrary to $\mathcal{L}^{>k}$, this bound puts as much emphasis on the lowest latent variables as the regular ELBO but keeps track of large deviation from the unconditional prior in the top $L-l$ KL-terms since it is not guided by the approximate posterior for $\z_{>l}$. We hypothesize that this bound might be useful for OOD detection in cases where the discriminating factor is to be found in low-level statistics rather than high-level features.

Additionally, we can incorporate it in a generalized log likelihood-ratio between $\mathcal{L}^{<l}$ and $\mathcal{L}^{>k}$
\begin{equation}
    LLR^{>k}_{<l} = \mathcal{L}^{<l} - \mathcal{L}^{>k}.
\end{equation}
We hypothesize that this score, or the other possible permutations of it, might be useful for OOD detection but leave further examination to future work.

\section{Note on the KL-term of hierarchical VAEs}
In this research we choose model parameterizations relying on bottom-up inference \cite{burda_importance_2016},
\begin{equation}
    q_\phi(\z|\x) = q_\phi(\z_1|\x)\textstyle\prod_{i=2}^{L} q_\phi(\z_i|\z_{i-1}) \ .
\end{equation}
We do this because bottom-up inference enables the model to learn covariance between the latent variables in the hierarchy.
In the inference model, any latent variable is dependent on the latent variables below it in the hierarchy and, importantly, the top most latent variable is dependent on all other latent variables.

In contrast, a top-down inference model \cite{sonderby_ladder_2016} has a topmost latent variable $\z_L$ that is independent of the other latent variables and is directly given by $\x$.
\begin{equation}
    q_\phi(\z|\x) = q_\phi(\z_L|\x)\textstyle\prod_{i=L-1}^{1} q_\phi(\z_{i}|\z_{i+1}) \ .
\end{equation}
This, in essence, makes $\z_L$ a mean-field approximation without any covariance structure tying it to the other latent variables, $\text{Cov}(z_{L,i}, z_{k,j})=0$ for $k<L$.
Furthermore, since the approximate posterior (and the prior) typically have diagonal covariance, $\z_L$ is also mean-field within its own elements, $\text{Cov}(z_{L,i}, z_{L,j})=0$ for $i\ne j$.

We hypothesize that the covariance of latent variables towards the top of the hierarchy with other latent variables is important for learning semantic representations.
However, top-down inference models are easier to optimize as has recently been demonstrated \cite{sonderby_ladder_2016, vahdat_nvae_2020, child_very_2021}.

In the following, we inspect the differences between the ELBO used for bottom-up inference and the ELBO used for top-down inference and show that it is not generally possible to decompose the total KL-divergence into separate KL-divergences per latent variable.
Specifically, for top-down inference it is possible to obtain KL-divergence at the top-most latent variable and an expectation of a KL-divergence for the other latent variables.
For bottom-up inference, the resulting terms are no longer KL-divergences except at the top-most latent variable.

We ask the question whether models relying on top-down inference are impeded in their use for semantic OOD detection, or whether they still learn to assign a more semantic representation in the top-most variables simply due to the flexibility of the deterministic neural network layers.
This remains an open research question.

\subsection{Bottom-up inference}
By splitting up the expectation, we can write the ELBO of a two-layer bottom-up hierarchical VAE as
\begin{align}
    \log p(\x) &\ge \mathbb{E}_{q(\z_1, \z_2|\x)} \left[ \log p(\x|\z_1) \right] \\
    &+ \mathbb{E}_{q(\z_1,\z_2|\x)} \left[ \log p(\z_1|\z_2) - \log q(\z_1|\x) \right] \notag \\
    &+ \mathbb{E}_{q(\z_1,\z_2|\x)} \left[ \log p(\z_2) - \log q(\z_2|\z_1) \right] \notag \ .
\end{align}
We can write out the expectations in order to derive the KL-divergence terms of the bottom-up ELBO:
\begin{align}
    \log p(\x) &\ge \int \int \log p(\x|\z_1) \text{d}\z_2\z_1 \label{eq:2 layer VAE bottom up ELBO integrals} \\
    &+ \int q(\z_1|\x) \int q(\z_2|\z_1) \log \frac{p(\z_1|\z_2)}{q(\z_1|\x)} \text{d}\z_2\z_1 \notag \\ 
    &+ \int q(\z_1|\x) \int q(\z_2|\z_1) \log \frac{p(\z_2)}{q(\z_2|\z_1)} \text{d}\z_2\z_1 \notag  \ .
\end{align}
From the above, we can see that since the decomposition is in a reverse order, we cannot derive the KL-divergence for the second term. This will hold in general for $L$-layered models for any latent variables  $\z_1,...,\z_{L-1}$:
\begin{align}
    \log p(\x) &\ge \mathbb{E}_{q(\z_1, \z_2|\x)} \left[ \log p(\x|\z_1) \right] \\
    &+ \mathbb{E}_{q(\z_1|\x)}\left[ \mathbb{E}_{q(\z_2|\z_1)}\left[\log \frac{p(\z_1|\z_2)}{q(\z_1|\x)}\right]\right] \notag \\
    &+ \mathbb{E}_{q(\z_1|\x)}\left[-D_\text{KL}[q(\z_2|\z_1)||p(\z_2)]\right] \notag \ .
\end{align}

\subsection{Top-down inference}
By splitting up the expectation, we can write the ELBO of a two-layer top-down hierarchical VAE as
\begin{align}
    \log p(\x) &\ge \mathbb{E}_{q(\z_1, \z_2|\x)} \left[ \log p(\x|\z_1) \right] \\ 
    &+ \mathbb{E}_{q(\z_1,\z_2|\x)} \left[ \log p(\z_2|\x) - \log q(\z_2|\x) \right] \notag \\
    &+ \mathbb{E}_{q(\z_1,\z_2|\x)} \left[ \log p(\z_1|\z_2) - \log q(\z_1|\z_2) \right] \notag \ .
\end{align}
We can write out the expectations in order to derive the KL-divergence terms:
\begin{align}
    \log p(\x) &\ge \int \int \log p(\x|\z_1) d\z_1\z_2 \\
    &+ \int q(\z_2|\x) \log \frac{p(\z_2|\x)}{q(\z_2|\x)} d\z_2 \notag \\ 
    &+ \int q(\z_2|\x) \int q(\z_1|\z_2) \log \frac{p(\z_1|\z_2)}{q(\z_1|\z_2)} d\z_1\z_2 \notag \ .
\end{align}
The KL-divergence terms can now easily be computed by:
\begin{align}
    \log p(\x) &\ge \mathbb{E}_{q(\z_1, \z_2|\x)} \left[ \log p(\x|\z_1) \right] \\
    &- D_\text{KL}[q(\z_2|\x)||p(\z_2)] \notag \\ 
    &- \mathbb{E}_{q(\z_2|\x)}\left[D_\text{KL}[q(\z_1|\z_2)||p(\z_1|\z_2)\right] \notag \ .
\end{align}
Note that the KL-divergence in the second layer is not exact since it is dependent on the sample-noise from the layer below. An exact solution can only be derived if the latent variables $\mathbf{z}$ are all conditionally independent. However, this comes at the cost of not learning a covariance structure.

\section{Additional results}
We provide additional results for a model trained on FashionMNIST in \autoref{tab:additional-results-fashionmnist}, a model trained on MNIST in \autoref{tab:additional-results-mnist}, a model trained on CIFAR10 in \autoref{tab:additional-results-cifar} and a model trained on SVHN in \autoref{tab:additional-results-svhn}.

We note that while the likelihood is highly unreliable across the datasets, the proposed log likelihood-ratio score is consistent and always allows correct OOD detection with high AUROC$\uparrow$.

\begin{table}[t]
    \centering
    \resizebox{\columnwidth}{!}{%
    \begin{tabular}{llrrr}
        \toprule
         OOD dataset & Metric & AUROC$\uparrow$ & AUPRC$\uparrow$ & FPR80$\downarrow$ \\
         \midrule
         \multicolumn{5}{c}{\textbf{Trained on SVHN}} \\
         \midrule
CIFAR10            &  $L^{>0}$         &  $0.992$  &  $0.993$  &  $0.004$  \\
CIFAR10            &  $L^{>1}$         &  $0.988$  &  $0.990$  &  $0.002$  \\
CIFAR10            &  $L^{>2}$         &  $0.746$  &  $0.756$  &  $0.468$  \\
CIFAR10            &  $LLR^{>1}$       &  $0.939$  &  $0.950$  &  $0.052$  \\
\midrule
SVHN               &  $L^{>0}$         &  $0.599$  &  $0.587$  &  $0.702$  \\
SVHN               &  $L^{>1}$         &  $0.555$  &  $0.543$  &  $0.755$  \\
SVHN               &  $L^{>2}$         &  $0.403$  &  $0.431$  &  $0.869$  \\
SVHN               &  $LLR^{>1}$       &  $0.489$  &  $0.484$  &  $0.799$  \\
         \bottomrule
    \end{tabular}
    }
    \caption{Additional results for the HVAE model trained on SVHN. All results computed with 1000 importance samples.}
    \label{tab:additional-results-svhn}
\end{table}

\begin{table}[t]
    \centering
    \resizebox{\columnwidth}{!}{%
    \begin{tabular}{llrrr}
        \toprule
         OOD dataset & Metric & AUROC$\uparrow$ & AUPRC$\uparrow$ & FPR80$\downarrow$ \\
         \midrule
         \multicolumn{5}{c}{\textbf{Trained on CIFAR10}} \\
         \midrule
SVHN          &  $L^{>0}$    &  0.083  &  0.318  &  0.974  \\
SVHN          &  $L^{>1}$    &  0.097  &  0.320  &  0.972  \\
SVHN          &  $L^{>2}$    &  0.693  &  0.725  &  0.599  \\
SVHN          &  $LLR^{>2}$  &  0.811  &  0.837  &  0.394  \\
\midrule
CIFAR10       &  $L^{>0}$    &  0.485  &  0.488  &  0.817  \\
CIFAR10       &  $L^{>1}$    &  0.467  &  0.476  &  0.822  \\
CIFAR10       &  $L^{>2}$    &  0.411  &  0.433  &  0.869  \\
CIFAR10       &  $LLR^{>1}$  &  0.469  &  0.479  &  0.835  \\
         \bottomrule
    \end{tabular}
    }
    \caption{Additional results for the HVAE model trained on CIFAR10. All results computed with 1000 importance samples.}
    \label{tab:additional-results-cifar}
\end{table}

\begin{table}[t]
    \centering
    \resizebox{\columnwidth}{!}{%
    \begin{tabular}{llrrr}
        \toprule
         OOD dataset & Metric & AUROC$\uparrow$ & AUPRC$\uparrow$ & FPR80$\downarrow$ \\
         \midrule
         \multicolumn{5}{c}{\textbf{Trained on FashionMNIST}} \\
         \midrule
MNIST                    & $\mathcal{L}^{>0}$     &  0.268  &  0.363  &  0.882 \\
MNIST                    & $\mathcal{L}^{>1}$     &  0.593  &  0.591  &  0.658 \\
MNIST                    & $\mathcal{L}^{>2}$     &  0.712  &  0.750  &  0.548 \\
MNIST                    & $LLR^{>1}$             &  0.986  &  0.987  &  0.011 \\
\midrule
notMNIST                 &  $\mathcal{L}^{>0}$    &  0.916  &  0.932  &  0.116 \\
notMNIST                 &  $\mathcal{L}^{>1}$    &  0.983  &  0.986  &  0.000 \\
notMNIST                 &  $\mathcal{L}^{>2}$    &  0.997  &  0.997  &  0.000 \\
notMNIST                 &  $LLR^{>1}$            &  0.998  &  0.998  &  0.000 \\
\midrule
KMNIST                   &  $\mathcal{L}^{>0}$    &  0.690  &  0.694  &  0.554 \\
KMNIST                   &  $\mathcal{L}^{>1}$    &  0.835  &  0.863  &  0.359 \\
KMNIST                   &  $\mathcal{L}^{>2}$    &  0.844  &  0.875  &  0.339 \\
KMNIST                   &  $LLR^{>1}$            &  0.974  &  0.977  &  0.017 \\
\midrule
Omniglot28x28            &  $\mathcal{L}^{>0}$    &  0.898  &  0.837  &  0.166 \\
Omniglot28x28            &  $\mathcal{L}^{>1}$    &  0.991  &  0.989  &  0.011 \\
Omniglot28x28            &  $\mathcal{L}^{>2}$    &  1.000  &  1.000  &  0.000 \\
Omniglot28x28            &  $LLR^{>2}$            &  1.000  &  1.000  &  0.000 \\
\midrule
Omniglot28x28Inverted    &  $\mathcal{L}^{>0}$    &  0.261  &  0.361  &  0.879 \\
Omniglot28x28Inverted    &  $\mathcal{L}^{>1}$    &  0.450  &  0.431  &  0.709 \\
Omniglot28x28Inverted    &  $\mathcal{L}^{>2}$    &  0.557  &  0.574  &  0.678 \\
Omniglot28x28Inverted    &  $LLR^{>1}$            &  0.954  &  0.954  &  0.050 \\
\midrule
SmallNORB28x28           &  $\mathcal{L}^{>0}$    &  0.982  &  0.984  &  0.000 \\
SmallNORB28x28           &  $\mathcal{L}^{>1}$    &  0.998  &  0.998  &  0.000 \\
SmallNORB28x28           &  $\mathcal{L}^{>2}$    &  1.000  &  1.000  &  0.000 \\
SmallNORB28x28           &  $LLR^{>2}$            &  0.999  &  0.999  &  0.002 \\
\midrule
SmallNORB28x28Inverted   &  $\mathcal{L}^{>0}$    &  0.965  &  0.971  &  0.000 \\
SmallNORB28x28Inverted   &  $\mathcal{L}^{>1}$    &  0.997  &  0.992  &  0.000 \\
SmallNORB28x28Inverted   &  $\mathcal{L}^{>2}$    &  0.981  &  0.985  &  0.000 \\
SmallNORB28x28Inverted   &  $LLR^{>2}$            &  0.941  &  0.946  &  0.069 \\
\midrule
FashionMNIST             &  $\mathcal{L}^{>0}$    &  0.476  &  0.484  &  0.816 \\
FashionMNIST             &  $\mathcal{L}^{>1}$    &  0.475  &  0.482  &  0.817 \\
FashionMNIST             &  $\mathcal{L}^{>2}$    &  0.475  &  0.484  &  0.823 \\
FashionMNIST             &  $LLR^{>1}$            &  0.488  &  0.496  &  0.811 \\
         \bottomrule
    \end{tabular}
    }
    \caption{Additional results for the HVAE model trained on FashionMNIST. All results computed with 1000 importance samples.}
    \label{tab:additional-results-fashionmnist}
\end{table}

\begin{table}[t]
    \centering
    \resizebox{\columnwidth}{!}{%
    \begin{tabular}{llrrr}
        \toprule
         OOD dataset & Metric & AUROC$\uparrow$ & AUPRC$\uparrow$ & FPR80$\downarrow$ \\
         \midrule
         \multicolumn{5}{c}{\textbf{Trained on MNIST}} \\
         \midrule
FashionMNIST                     &  $L^{>0}$  &  $1.000$  &  $1.000$  &  $0.000$ \\
FashionMNIST                     &  $L^{>1}$  &  $1.000$  &  $1.000$  &  $0.000$ \\
FashionMNIST                     &  $L^{>2}$  &  $0.981$  &  $0.983$  &  $0.003$ \\
FashionMNIST                   &  $LLR^{>1}$  &  $0.999$  &  $0.999$  &  $0.000$ \\
\midrule
notMNIST                         &  $L^{>0}$  &  $1.000$  &  $1.000$  &  $0.000$ \\
notMNIST                         &  $L^{>1}$  &  $1.000$  &  $1.000$  &  $0.000$ \\
notMNIST                         &  $L^{>2}$  &  $1.000$  &  $1.000$  &  $0.000$ \\
notMNIST                       &  $LLR^{>1}$  &  $1.000$  &  $0.999$  &  $0.000$ \\
\midrule
KMNIST                           &  $L^{>0}$  &  $1.000$  &  $1.000$  &  $0.000$ \\
KMNIST                           &  $L^{>1}$  &  $1.000$  &  $1.000$  &  $0.000$ \\
KMNIST                           &  $L^{>2}$  &  $0.987$  &  $0.987$  &  $0.011$ \\
KMNIST                         &  $LLR^{>1}$  &  $0.999$  &  $0.999$  &  $0.000$ \\
\midrule
Omniglot28x28                    &  $L^{>0}$  &  $1.000$  &  $1.000$  &  $0.000$ \\
Omniglot28x28                    &  $L^{>1}$  &  $1.000$  &  $1.000$  &  $0.000$ \\
Omniglot28x28                    &  $L^{>2}$  &  $1.000$  &  $1.000$  &  $0.000$ \\
Omniglot28x28                  &  $LLR^{>1}$  &  $1.000$  &  $1.000$  &  $0.000$ \\
\midrule
Omniglot28x28Inverted            &  $L^{>0}$  &  $0.862$  &  $0.902$  &  $0.205$ \\
Omniglot28x28Inverted            &  $L^{>1}$  &  $0.923$  &  $0.943$  &  $0.056$ \\
Omniglot28x28Inverted            &  $L^{>2}$  &  $0.749$  &  $0.691$  &  $0.411$ \\
Omniglot28x28Inverted          &  $LLR^{>1}$  &  $0.944$  &  $0.953$  &  $0.057$ \\
\midrule
SmallNORB28x28                   &  $L^{>0}$  &  $1.000$  &  $1.000$  &  $0.000$ \\
SmallNORB28x28                   &  $L^{>1}$  &  $1.000$  &  $1.000$  &  $0.000$ \\
SmallNORB28x28                   &  $L^{>2}$  &  $1.000$  &  $1.000$  &  $0.000$ \\
SmallNORB28x28                 &  $LLR^{>1}$  &  $1.000$  &  $1.000$  &  $0.000$ \\
\midrule
SmallNORB28x28Inverted           &  $L^{>0}$  &  $1.000$  &  $1.000$  &  $0.000$ \\
SmallNORB28x28Inverted           &  $L^{>1}$  &  $1.000$  &  $1.000$  &  $0.000$ \\
SmallNORB28x28Inverted           &  $L^{>2}$  &  $0.977$  &  $0.980$  &  $0.001$ \\
SmallNORB28x28Inverted         &  $LLR^{>1}$  &  $0.985$  &  $0.987$  &  $0.000$ \\
\midrule
MNIST                            &  $L^{>0}$  &  $0.488$  &  $0.486$  &  $0.807$ \\
MNIST                            &  $L^{>1}$  &  $0.469$  &  $0.469$  &  $0.816$ \\
MNIST                            &  $L^{>2}$  &  $0.514$  &  $0.505$  &  $0.791$ \\
MNIST                          &  $LLR^{>2}$  &  $0.515$  &  $0.507$  &  $0.792$ \\
         \bottomrule
    \end{tabular}
    }
    \caption{Additional results for the HVAE model trained on MNIST. All results computed with 1000 importance samples.}
    \label{tab:additional-results-mnist}
\end{table}

\end{document}